\documentclass{article}

\usepackage[main, final]{neurips_2026}

\usepackage[table,dvipsnames,svgnames]{xcolor}
\usepackage[utf8]{inputenc} 
\usepackage[T1]{fontenc}    
\usepackage{hyperref}       
\usepackage{url}            
\usepackage{booktabs}       
\usepackage{amsfonts}       
\usepackage{nicefrac}       
\usepackage{microtype}      
\usepackage{titletoc}
\newcommand\DoToC{%
  \startcontents
  \printcontents{}{1}{\hrulefill\vskip0pt}
  \vskip0pt \noindent\hrulefill
  }
\usepackage[dvipsnames]{xcolor}
\usepackage[svgnames]{xcolor}
\makeatletter
    \newcommand{\colorboxed}[3][boxcolor]{\fcolorbox{#2}{#1}{\m@th$\displaystyle#3$}}
\makeatother

\usepackage{amsmath}
\usepackage{amssymb}
\usepackage{mathtools}
\usepackage{amsthm}
\usepackage{setspace}
\usepackage{ltablex,booktabs,ragged2e,caption}
\usepackage[capitalize,noabbrev]{cleveref}
\usepackage{wrapfig}
\usepackage{url}
\usepackage{algorithm, algorithmic}
\usepackage{graphicx}
\usepackage{multirow}
\usepackage{dsfont}
\usepackage{enumitem}
\usepackage[table]{xcolor}
\usepackage{adjustbox}
\hypersetup{
    colorlinks,
    linkcolor={RoyalBlue},
    citecolor={RoyalBlue},
    urlcolor={RoyalBlue}
}
\definecolor{Gray}{gray}{0.93}
\definecolor{lightgray}{gray}{0.97}
\definecolor{LightWheat}{HTML}{F9F3E5}
\definecolor{softgreen}{RGB}{228,255,246}
\definecolor{brightgreen}{RGB}{227,255,215}
\definecolor{softpurple}{RGB}{221,194,255}
\definecolor{purpleblue}{RGB}{205,211,255}
\definecolor{softpink}{RGB}{255,220,241}
\definecolor{brown}{RGB}{254,255,121}
\definecolor{orrange}{RGB}{255,138,69}
\usepackage[most]{tcolorbox}
\usepackage{minitoc}
\newtcolorbox{mybox}[1][]{
enhanced jigsaw,
colback=LightWheat,
boxrule=0pt,
overlay unbroken and first ={},
breakable,
top=1pt,bottom=1pt,
left=1pt,right=1pt,
arc=0pt,outer arc=0pt,
#1}
\theoremstyle{plain}
\newtheorem{theorem}{Theorem}[section]
\newtheorem{proposition}[theorem]{Proposition}

\theoremstyle{definition}
\newtheorem{definition}[theorem]{Definition}
\newtheorem{assumption}[theorem]{Assumption}
\theoremstyle{remark}

\DeclareMathOperator*{\argminA}{arg\,min}
\title{Causal Heteroscedastic Structure Learning From Incomplete Temporal data}

\author{%
  Abdellah Rahmani \\
  LTS4, EPFL\\
  Lausanne, Switzerland \\
  \texttt{abdellah.rahmani@epfl.ch} \\
   \And
  Pascal Frossard \\
  LTS4, EPFL\\
  Lausanne, Switzerland \\
  \texttt{pascal.frossard@epfl.ch} \\
}

\begin{document}

\maketitle

\begin{abstract}
Discovering causal structures in multivariate time series (MTS) is critical in domains such as finance and neuroscience, yet remains challenging in practice due to three distinct obstacles. First, MTS are routinely corrupted by missing values, where naive impute-then-discover pipelines distort the underlying data distribution and yield spurious causal graphs. Second, identifying instantaneous effects is difficult even on complete data, and missingness further entangles imputation with graph recovery. Third, real-world data frequently exhibits heteroscedastic noise with variance driven by both instantaneous and lagged causes, violating the core assumptions of standard discovery methods. Existing approaches fail to address these challenges concurrently, typically presuming complete data, homoscedastic noise, or strictly time-lagged relationships. Crucially, we show through ablation studies that resolving these issues sequentially, rather than jointly, leads to a substantial degradation in performance. To overcome this, we introduce Cheesefill, an optimal transport framework for causal structure learning under missingness and heteroscedastic noise. Cheesefill parameterizes causal mechanisms using conditional normalizing flows and jointly learns a stochastic correction map that refines naive imputations into trajectories consistent with the inferred causal mechanisms. Theoretically, we prove that optimizing over stochastic correction maps is equivalent to a conditional kernel reformulation of the Kantorovich problem, from which our neural parameterization yields a tractable objective. Extensive experiments on synthetic and real-world benchmarks demonstrate that Cheesefill consistently outperforms existing baselines across diverse missingness regimes.

\end{abstract}
\section{Introduction}
Causal structure learning from multivariate time series (MTS) is a fundamental problem with applications spanning traffic modeling \citep{cheng2023causaltime}, biology \citep{sachs2005causal}, climate science \citep{runge2019detecting, rahmanicausal}, and healthcare \citep{shen2020challenges}. Yet recovering causal relationships from real-world MTS confronts three intertwined challenges. First, observations are routinely corrupted by missing values \citep{white2011multiple}, typically due to sensor failures or transmission loss; robustness to such missingness is essential but remains underexplored. Second, when the sampling interval is coarse relative to the underlying dynamics, multiple events collapse onto the same timestamp, giving rise to instantaneous causal links that any faithful causal model must capture. Third, real MTS exhibit heteroscedastic noise \citep{gunther2022conditional} whose variance depends on both instantaneous and lagged causes, a phenomenon documented in fMRI \citep{seiler2017multivariate}, EEG \citep{huizenga1995equivalent, rahmani2023meta}, and financial data \citep{hamilton1994autoregressive}.

Most existing causal discovery methods for MTS assume complete observations and homoscedastic noise. On complete MTS, the field has developed in three directions: Granger-based neural 
approaches \citep{lowe2022amortized, bussmann2021neural} that infer dependencies through component-wise prediction; score-based formulations such as Dynotears \citep{pamfil2020Dynotears}, which recovers instantaneous and lagged links via the differentiable acyclicity constraint introduced by Zheng et al. \citep{zheng2018dags}; and constraint-based methods such as PCMCI+ \citep{runge2020discovering}, which scale conditional independence testing to high-dimensional time series with both lagged and 
contemporaneous links. Closer to our setting, Gong et al. \citep{gong2022rhino} consider structural equation models with historically dependent noise, and Rahmani et al. \citep{rahmaniflow} handle heteroscedasticity in stationary and non-stationary regimes, but again on complete MTS. When data is missing, the prevailing approach is a two-stage pipeline: fill the gaps with zero padding, interpolation, or learned imputer \citep{cini2021filling, cao2018brits, wang2023timemixer}, then application of a causal discovery algorithm. However, recent work \citep{cheng2023cuts, cheng2024cuts+, vo2024optimal} shows that lower reconstruction error does not translate into better recovered graphs: the proxy complete distribution induced by imputation encodes a different set of conditional independencies, and may therefore be compatible with a causal structure that is distinct from the actual ground truth. This calls for a joint, end-to-end treatment of imputation and structure learning. Existing methods in this direction \citep{cheng2023cuts,cheng2024cuts+} however assume purely time-lagged relationships and homoscedastic noise, leaving the simultaneous treatment of instantaneous links, heteroscedasticity, and missingness as an open problem. A detailed related work appears in  Appendix~\ref{app:related_work}.
\begin{figure}[t]
    \centering
    \includegraphics[width=1\linewidth]{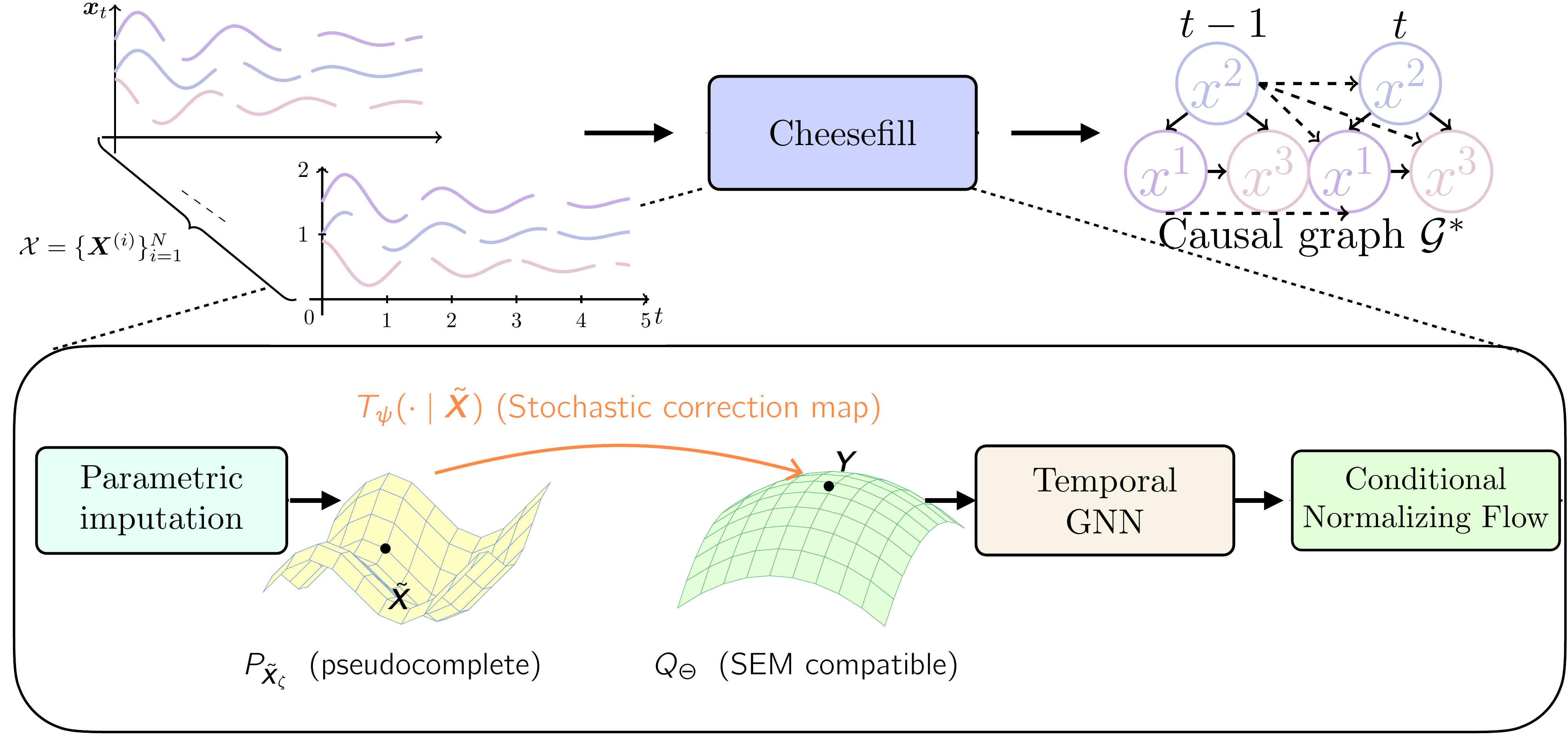}
    \caption{Cheesefill overview. The incomplete MTS dataset 
$\mathcal{X}$ is first processed by a parametric imputer with parameters $\zeta$. 
A stochastic correction map  $T_\psi$ refines $\tilde{\boldsymbol{X}}$ into a sample $\boldsymbol{Y}$ consistent with the structural equation model marginal $Q_\Theta$. TGNN and CNF, models causal mechanisms and heteroscedastic noise. Joint optimisation over $(\Theta, \psi, \zeta)$ recovers the temporal causal graph 
$\mathcal{G}^*$.
}
    \label{fig:fantom_over}
\end{figure}

To solve the three challenges, we propose Cheesefill (illustrated in Figure \ref{fig:fantom_over}), an optimal transport (OT) framework \citep{muzellec2020missing} for causal structure learning from MTS with instantaneous and lagged effects, heteroscedastic noise \citep{rahmaniflow}, and missing values. To the best of our knowledge, Cheesefill is the first method to recover instantaneous and time-lagged links from incomplete MTS, with or without heteroscedastic noise. Given an MTS with missing entries, Cheesefill couples Bayesian structure learning \citep{geffner2022deep, rahmaniflow, gong2022rhino}, which filters out 
spurious edges, with a conditional normalizing flow \citep{durkan2019neural} that models heteroscedastic noise, while simultaneously learning a stochastic correction map that refines naive imputations into samples consistent with the causal model. On the theoretical side, we show that optimizing over admissible stochastic correction maps is equivalent to a conditional kernel reformulation of the Kantorovich problem \citep{kantorovich2006problem}; the neural parameterization used by Cheesefill then yields a tractable objective. On the experimental side, Cheesefill consistently outperforms causal discovery methods designed for incomplete MTS and naive \textit{impute then discover} pipelines on synthetic benchmarks and two real world datasets. Our contributions are:
\begin{itemize}
    \item We introduce Cheesefill, a framework for recovering instantaneous and time lagged causal links from incomplete MTS under homoscedastic or heteroscedastic noise.
    \item We cast the problem as OT-based density fitting, combining Bayesian structure learning, conditional normalizing flows for heteroscedastic mechanisms, and a stochastic correction map for aligning naive imputations with the causal model.
    \item We show that admissible stochastic correction maps give a conditional-kernel reformulation of the Kantorovich problem; Cheesefill’s neural correction map yields a tractable objective.
    \item We demonstrate consistent gains over different baselines across synthetic and real datasets under multiple missingness mechanisms.
\end{itemize}

\section{Preliminaries}\vspace{-5pt}
\textbf{Notation. } We use calligraphic letters (e.g. $\mathcal{X}$) for sets, uppercase bold ($\boldsymbol{G}_{\tau}$) for matrices, lowercase bold ($\boldsymbol{x}_t$) for vectors, and lowercase letters ($x_{t-\tau}^i$) for scalars. Ground-truth quantities are marked with an asterisk, e.g. $\mathcal{G}^*$. We assume that all distributions $P(\boldsymbol{X})$ admit lower bounded densities $p(\boldsymbol{x}_{t})$ with respect to the Lebesgue measure. The notation $[|0: L|]$ denotes the integer set $\{0,\dots,L\}$, and $|\cdot|$ denotes set cardinality. A MTS is written $\boldsymbol{X} = (\boldsymbol{x}_t)_{t \in \mathcal{T}} = (x_t^i)_{i \in \mathbf{V},\, t \in \mathcal{T}}$, with $|\mathbf{V}| = d$ components and length $|\mathcal{T}|$, where 
$\mathcal{T}$ is the time index set. We write $\boldsymbol{x}_{<t}$ for the history 
$\{\boldsymbol{x}_{t-L},\dots,\boldsymbol{x}_{t-1}\}$.

\textbf{Optimal Transport (OT).}
A central object in our framework is the cost of transforming one distribution
into another. Kantorovich's formulation \citep{kantorovich2006problem} of this
problem reads
\begin{equation}
W_c(P, Q)
:=
\inf_{\Gamma \in \mathcal{P}(X \sim P,\, Y \sim Q)}
\mathbb{E}_{(X, Y) \sim \Gamma}\bigl[c(X, Y)\bigr],
\label{defkantorovich}
\end{equation}
where $c : \Omega_X \times \Omega_X \to \mathbb{R}_+$ is any measurable cost function on $(\Omega_X, \mathcal{B}(\Omega_X))$, the measurable space of complete length-$|\mathcal{T}|$, $d$-variate MTS trajectories, which we assume to be a standard Borel space, and $\mathcal{P}(X \sim P, Y \sim Q)$ denotes the set of all couplings of $(X, Y)$ with marginals $P$ and $Q$. We will also use an equivalent conditional representation of couplings, defined as follows:
\begin{mybox}
\begin{definition}[Admissible stochastic transport kernel]
\label{def:admissible_kernel}
A stochastic kernel
\(T(dy\mid x)\) from \(\Omega_X\) to \(\Omega_X\) is admissible between
\(P\) and \(Q\) if, for every measurable set \(A\in\mathcal B_X\),
\[
\int_{\Omega_X} T(A\mid x)\,dP(x)
=
Q(A).
\]
We denote the set of admissible kernels between \(P\) and \(Q\) by
\(\mathcal{F}(P,Q)\).
\end{definition}
\end{mybox}
On the finite dimensional trajectory spaces, \textcolor{RoyalBlue}{as considered in this paper}, every coupling
\(\Gamma\in\mathcal{P}(X\sim P,\,Y\sim Q)\) admits a conditional representation
\[
\Gamma(dx,dy)
=
P(dx)T_\Gamma(dy\mid x).
\]
We now introduce the object our method recovers. Following \citep{assaad2022survey,rahmaniflow, gong2022rhino}, we represent the causal structure of an MTS as a collection of lag-specific adjacency matrices, which separates instantaneous links ($\tau = 0$) from lagged ones ($\tau > 0$).
\begin{mybox}\begin{definition}[Temporal Causal Graph 
\citep{rahmaniflow, rahmanicausal}]
The temporal causal graph associated with an MTS $\boldsymbol{X} = (\boldsymbol{x}_t)_{t \in \mathcal{T}}$ is a Directed Acyclic Graph (DAG) $\mathcal{G}=(\mathbf{V}, E)$, 
represented by a collection of adjacency matrices $\{\boldsymbol{G}_{\tau}\}_{\tau \in [|0: L|]} = \{\boldsymbol{G}_{0},\dots,\boldsymbol{G}_{L}\}$, with a fixed maximum lag $L$. Its vertices $\mathbf{V}$ consist of the components $x_{t'}^1, \ldots, x_{t'}^d$ for each $t' \in [|t-L:t|]$. For $\tau \in [|1:L|]$, the variables $x_{t-\tau}^i$ and $x_t^j$ are connected by a lag-specific directed link $x_{t-\tau}^i \rightarrow x_t^j$ in $\mathcal{G}$ if and only 
if $x^i$ at time $t-\tau$ causes $x^j$ at time $t$; in that case $[G_{\tau}]_{ij}$ is non-zero and $x^i \in \mathbf{Pa}_j^{<t}$, the set of lagged parents of $j$. For instantaneous links ($\tau = 0$), self-loops are excluded ($i \neq j$): we have an edge $x_{t}^i \rightarrow x_t^j$ with $x^i \in \mathbf{Pa}_j^{t}$, the set of instantaneous parents of $j$ at time $t$, if and only if $x^i$ at time $t$ causes $x^j$ at time $t$.
\label{definitionn1}
\end{definition}\end{mybox}

\textbf{Structural equation model (SEM). } Let $\mathcal{X}=\bigl\{\boldsymbol{X}^{(1)},\cdots, \boldsymbol{X}^{(N)}\bigr\}$ be a collection of $N$ MTS, where each $\boldsymbol{X}^{(j)}$ has length $|\mathcal{T}|$ and consists of $|\mathbf{V}| = d$ variables $\{x_{t}^{i,(j)}\}_{t \in \mathcal{T},\, 
i \in \mathbf{V}}$. We assume that all $\boldsymbol{X}^{(j)} \in \mathcal{X}$ share the same temporal causal 
graph $\mathcal{G}$ with maximum lag $L$ as in Definition~\ref{definitionn1}. The observation $\boldsymbol{x}_t^{(j)}$ then follows the general SEM
\begin{equation}
x_{t}^{i,(j)} 
= f^{i}\!\bigl(\mathbf{Pa}_i^{<t},\, \mathbf{Pa}_i^{t}\bigr) 
+ g^{i}\!\bigl(\mathbf{Pa}_i^{<t},\, \mathbf{Pa}_i^{t}\bigr) 
\cdot \epsilon_t^{i,(j)},
\label{eq_sem_hetero}
\end{equation}
where $f^{i}$ and $g^{i}$ are general differentiable functions with $g^{i}$ strictly positive, and $\epsilon_t^{i,(j)}$ follows an arbitrary density $p_{\epsilon}$ with $\mathbb{E}(\epsilon_t^{i,(j)}) = 0$ and 
$\mathbb{E}\bigl((\epsilon_t^{i,(j)})^2\bigr) = 1$ (without loss of generality). This single SEM covers both regimes that interest us: when $g^{i}$ is constant we recover the homoscedastic case, and when $g^{i}$ depends on the parents we capture the heteroscedastic noise observed in fMRI, EEG.

\textbf{SEM-induced mechanism density. } \label{mechanismdensity} Eq.~\eqref{eq_sem_hetero} not only specifies how to sample 
$x_t^{i}$, it also induces a density on the value $x_t^{i}$ once the parent values are fixed. For any candidate parent values $\mathbf{p}^{<t} \in \mathbb{R}^{|\mathbf{Pa}_i^{<t}|}$ and $\mathbf{p}^{t} \in \mathbb{R}^{|\mathbf{Pa}_i^{t}|}$, the random variable $\textcolor{teal}{f^{i}(\mathbf{p}^{<t} , \mathbf{p}^{t}) 
+ g^{i}(\mathbf{p}^{<t} , \mathbf{p}^{t})  \cdot \epsilon}$ admits a density on $\mathbb{R}$ which we denote $\textcolor{teal}{\rho^{i}\bigl(x \,;\, \mathbf{p}^{<t} , \mathbf{p}^{t}, \mathcal{G}\bigr)}$ with $\mathbb{E}[\rho^{i}\bigl(\cdot \,;\, \mathbf{p}^{<t} , \mathbf{p}^{t}, \mathcal{G}\bigr)] = f^{i}(\mathbf{p}^{<t} , \mathbf{p}^{t})$ and 
$\mathbb{E}\bigl[(\rho^{i}\bigl(\cdot \,;\, \mathbf{p}^{<t} , \mathbf{p}^{t}, \mathcal{G}\bigr) - f^{i}(\mathbf{p}^{<t} , \mathbf{p}^{t}))^2\bigr] = g^{i}(\mathbf{p}^{<t} , \mathbf{p}^{t})^2$ , 
with the dependence on $\mathcal{G}$ entering through the parent sets $\mathbf{Pa}_i^{<t},\, \mathbf{Pa}_i^{t}$ used to read off $\mathbf{p}^{<t} , \mathbf{p}^{t}$.

We use $\rho^{i}$ throughout the paper as a \textcolor{teal}{\emph{mechanism level}} density that scores observations against the SEM: it answers the question ``\textcolor{RoyalBlue}{\emph{given that the parents took these values, how plausible is the observation $x$? }}''. What makes $\rho^{i}$ central to our setting is the fact that it is agnostic to the source of the parent values, they may come from the data, from a model roll-out, or from an imputed sample, and the density remains the same family indexed by those inputs. This locality is what makes $\rho^{i}$ usable under missingness, where parent values must themselves be inferred.

\textbf{Missingness. } In practice, the $N$ MTS $\mathcal{X}=\bigl\{\boldsymbol{X}^{(1)},\cdots, 
\boldsymbol{X}^{(N)}\bigr\}$ we observe are rarely complete. We model missingness through a binary indicator matrix $\boldsymbol{M}^{(j)} \in \{0,1\}^{|\mathcal{T}|\times d}$, where 
$\boldsymbol{M}_t^{i,(j)}=1$ marks the $i^{\text{th}}$ feature at timestamp $t$ of MTS $(j)$ as missing and $\boldsymbol{M}_t^{i,(j)}=0$ marks it as observed. Missing entries are initially set to NaN, and they arise from a range of practical causes including entry errors, deliberate non responses, and sampling drops \citep{little2019statistical}. Following Little and Rubin \citep{little2019statistical}, missingness is classified into three regimes: \emph{missing completely at random} (MCAR), where the cause is independent of the data; \emph{missing at random} (MAR), where the probability of being missing depends only on observed values; and \emph{missing not at random} (MNAR), where the cause itself is unobserved. 
\begin{mybox}
\textbf{Identifiability. }Throughout the paper, and following \citep{gong2022rhino, rahmanicausal, rahmaniflow}, we use the standard assumptions of temporal causal discovery: the causal Markov property, stationarity, minimality, and causal sufficiency; see Appendix~\ref{app_assumptions} for precise statements. Under these assumptions, temporal SEMs with heteroscedastic noise have been shown to be uniquely identifiable from the observed distribution \citep{rahmaniflow}. For MCAR and MAR data, the missingness process is ignorable and the parameters can be consistently estimated based only on the observed data under large sample assumption\citep{little2019statistical}.
\end{mybox}

MNAR data is fundamentally more challenging because the observation process may depend on unobserved values and is therefore generally non ignorable. In the fully general case, the joint distribution is not recoverable without additional information; existing works use a missingness graph to establish graphical conditions for recovery \citep{bhattacharya2020, nabi2020, ma2021}. Such structural knowledge is difficult to obtain in practice, especially in our setting where the causal graph is also unknown. We therefore find it crucial to develop a principled approach that does not assume access to the underlying missingness mechanism. While we do not claim arbitrary MNAR identifiability without assumptions on the observation process, we provide extensive empirical evidence in Appendix~\ref{app_mnarresults} showing that Cheesefill remains effective in MNAR cases.
\vspace{-5pt}\section{Problem formulation}
\vspace{-5pt}
Let \((\Omega_X,\mathcal B_X)\) denote the measurable space of complete length \(|\mathcal T|\), \(d\) variate MTS trajectories, and let
\(P_{\boldsymbol X}\) \textcolor{teal}{be the unknown complete data distribution on}
\(\Omega_X\). We observe a finite collection of incomplete MTS $\mathcal{X}=\bigl\{\boldsymbol{X}^{(1)},\cdots, \boldsymbol{X}^{(N)}\bigr\}$ with associated masks $\mathcal{M}=\bigl\{\boldsymbol{M}^{(1)},\cdots, \boldsymbol{M}^{(N)}\bigr\}$. We assume that each latent complete trajectory
\(\boldsymbol X_\star^{(j)}\in\Omega_X\) is drawn i.i.d. from \(P_{\boldsymbol X}\), generated by the SEM in
Eq.~\eqref{eq_sem_hetero} under a shared \(\mathcal G^\ast\). The observed incomplete MTS \(\boldsymbol X^{(j)}\) is obtained by corrupting \(\boldsymbol X_\star^{(j)}\) according to the mask \(\boldsymbol M^{(j)}\), under MCAR, MAR, or MNAR missingness. Given the finite dataset \((\mathcal X,\mathcal M)\), our goal is to jointly recover \(\mathcal G^\ast\) and the mechanism parameters \(\Theta\) of \((f^i,g^i)_{i\in\mathbf V}\).
\vspace{-5pt}
\subsection{Idealised density-fitting formulation}
\label{probform} \vspace{-5pt}
\textbf{OT motivation. }We frame causal recovery as a density-fitting problem using the Kantorovich OT formulation. Unlike complete-data causal discovery, where the SEM can be scored directly on observed trajectories, the complete trajectories are unknown under missingness. Directly maximizing a score on fixed imputations would make the learned graph depend on artifacts of the imputer. The OT formulation instead compares the pseudo complete distribution and the SEM induced distribution, providing a principled way to couple imputation and graph learning through mechanism compatibility rather than treating imputation as a separate preprocessing step.\\
In the \textcolor{teal}{idealised complete data setting}, we would seek $(\Theta, \mathcal{G})$ 
that minimise the OT cost between the model distribution $Q_{\Theta}$ and the data distribution $P_{\boldsymbol{X}}$,
\begin{equation}
\Theta^*, \mathcal{G}^* \;=\; 
\argminA_{\Theta,\mathcal{G}}\;
W_{c_{\mathcal{G}}}(P_{\boldsymbol{X}}, Q_{\Theta}) \quad\text{subject to } \mathbf{G}_0 \in \text{DAGs},
\label{eq:ot_cost}
\end{equation}
where $W_{c_{\mathcal{G}}}$ is the Kantorovich cost in Eq.~\eqref{defkantorovich}. The choice of transport cost is crucial. A geometric cost such as \(\|X-Y\|^2\) measures pointwise proximity between trajectories, but causal structure is identified through local conditional mechanisms, not through
ambient Euclidean distance. We therefore use the SEM mechanism as the transport cost: $c_{\mathcal{G}}(\boldsymbol{X}, \boldsymbol{Y}) 
= -\sum_{t=1}^{|\mathcal{T}|}\sum_{i=1}^{d}\log\rho^{i}\bigl(x \,;\, \mathbf{p}^{<t} , \mathbf{p}^{t}, \mathcal{G}\bigr). $

For a fixed candidate parent trajectory \(\boldsymbol{Y}\), this cost measures how plausible the child values in \(\boldsymbol{X}\) are under the mechanisms specified by graph \(G\). Thus, low transport cost requires not only distributional proximity, but compatibility with the causal mechanisms of the candidate SEM. Eq.~\eqref{eq:ot_cost} cannot be solved directly in our setting for two reasons: (i) we do not have access to samples from $P_{\boldsymbol{X}}$, only to the corrupted collection $(\mathcal{X}, \mathcal{M})$; and (ii) defining a model distribution $Q_\Theta$ over incomplete MTS is not straightforward. Sections~\ref{sec:correction_map_formulation} and \ref{cheesefillsection} introduce Cheesefill, which resolves both through a stochastic correction map and a tractable parametric objective that can be optimized jointly with the causal graph and mechanisms.
\vspace{-5pt}
\subsection{Tractable density-fitting formulation} 
\label{sec:correction_map_formulation}
\vspace{-5pt}
Cheesefill addresses the aforementioned obstacles in section \ref{probform} in two steps. First, a parametric initializer converts the incomplete observations into pseudo complete trajectories, giving a tractable source distribution. Second, a stochastic correction map transports these pseudo complete trajectories toward refined trajectories used by the SEM mechanisms. 
\begin{mybox}
\paragraph{Intuition. }Recovering causal structure from incomplete MTS does not require necessarily reconstructing the full observational marginal \(P_{\boldsymbol X}\). In our setting, the graph is identified through mechanism-level constraints. Thus, \textcolor{RoyalBlue}{the role of imputation is} not merely to minimize reconstruction error, but \textcolor{RoyalBlue}{to produce completed trajectories whose conditional mechanisms are compatible with those implied by the candidate SEM}. \textit{Cheesefill implements this principle in two stages}. A parametric initializer first produces regular shaped pseudo complete trajectories. A stochastic correction map then adjusts these trajectories toward completions that both preserve the observed entries and better satisfy the local mechanism densities \(\rho_\Theta^i\). 
\end{mybox}
\paragraph{Parametric imputer.}
Cheesefill first applies a parametric imputer (PI) \(h_\zeta\) to fill the missing entries
of each MTS, yielding pseudo complete MTS  
$
\tilde{\mathcal X}_\zeta
=
\{\tilde{\boldsymbol X}^{(1)},\ldots,\tilde{\boldsymbol X}^{(N)}\},
\tilde{\boldsymbol X}\sim P_{\tilde{\boldsymbol X}_\zeta}.
$
Thus, \(P_{\tilde{\boldsymbol X}_\zeta}\) is the PI induced distribution on the complete MTS trajectory space \(\Omega_X\), providing a tractable source in place of the unavailable complete data distribution \(P_{\boldsymbol X}\) in Eq.~\eqref{eq:ot_cost}. Directly minimizing \(W_{c_{\mathcal G}}(P_{\tilde{\boldsymbol X}_\zeta},Q_\Theta)\) can inherit the failure mode of impute then discover pipelines \citep{cheng2023cuts, cheng2024cuts+}. 

\paragraph{Stochastic correction map.}
The pseudo complete distribution \(P_{\tilde{\boldsymbol X}_{\zeta}}\) gives a tractable source distribution, but its samples may contain dependencies induced by the imputer rather than by the underlying SEM. Cheesefill therefore does not treat pseudo complete trajectories as fixed causal evidence. Instead, Cheesefill introduces a \emph{stochastic correction map} \(T_{\psi}(d\boldsymbol Y\mid \tilde{\boldsymbol X})\), which transports each pseudo complete trajectory \(\tilde{\boldsymbol X}\) to a refined trajectory whose values instantiate the instantaneous and lagged parents in the mechanism cost \(c_{\mathcal G}(\tilde{\boldsymbol X},\boldsymbol Y)\). Thus, the following proposition shows that optimizing over such stochastic correction maps is not merely a heuristic relaxation: in the unrestricted case, it is exactly a conditional kernel reformulation of the Kantorovich problem.
\begin{mybox}\begin{proposition}[]
\label{thm:correction_ot_equivalence}
Let \((\Omega_X,\mathcal B(\Omega_X))\) be the measurable space of complete
length \(|\mathcal{T}|\), \(d\) variate MTS trajectories. Assume that \(\Omega_X\) is a
standard Borel space. Let \(P_{\tilde{\boldsymbol X}_{\zeta}}\) be the
pseudo complete data distribution on \(\Omega_X\), and let \(Q_\Theta\) be the
SEM induced complete trajectory distribution on \(\Omega_X\). Let \(c_{\mathcal G}:\Omega_X\times\Omega_X\to\mathbb R_+\) be a measurable
mechanism cost satisfying $\mathbb E_{(X,Y)\sim\Gamma}\bigl[|c_{\mathcal G}(X,Y)|\bigr]<\infty$
for all couplings \(\Gamma\) considered. Then
\begin{equation}
W_{c_{\mathcal G}}
\bigl(
P_{\tilde{\boldsymbol X}_{\zeta}},
Q_\Theta
\bigr)
=
\inf_{T\in\mathcal F(P_{\tilde{\boldsymbol X}_{\zeta}},Q_\Theta)}
\mathbb{E}_{\tilde{\boldsymbol X}\sim P_{\tilde{\boldsymbol X}_{\zeta}}}
\mathbb{E}_{\boldsymbol Y\sim T(\cdot\mid\tilde{\boldsymbol X})}
\bigl[
c_{\mathcal G}(\tilde{\boldsymbol X},\boldsymbol Y)
\bigr],
\label{eq:correction_ot_equivalence}
\end{equation}
where \(\mathcal F(P_{\tilde{\boldsymbol X}_{\zeta}},Q_\Theta)\) is the set of
admissible stochastic kernels in Definition~\ref{def:admissible_kernel}.
Moreover, for any parametric family \(\{T_\psi\}_{\psi\in\Psi}\), let $\Psi_{\mathrm{adm}}
=
\{\psi\in\Psi:
T_\psi\in\mathcal F(P_{\tilde{\boldsymbol X}_{\zeta}},Q_\Theta)\}.$ Then
\begin{equation}
W_{c_{\mathcal G}}
\bigl(
P_{\tilde{\boldsymbol X}_{\zeta}},
Q_\Theta
\bigr)
\leq
\inf_{\psi\in\Psi_{\mathrm{adm}}}
\mathbb{E}_{\tilde{\boldsymbol X}\sim P_{\tilde{\boldsymbol X}_{\zeta}}}
\mathbb{E}_{\boldsymbol Y\sim T_\psi(\cdot\mid\tilde{\boldsymbol X})}
\bigl[
c_{\mathcal G}(\tilde{\boldsymbol X},\boldsymbol Y)
\bigr].
\label{eq:parametric_upper_bound}
\end{equation}
The inequality is tight whenever the parametric family contains an optimal
admissible kernel. More generally, if the family contains an
\(\varepsilon\)-optimal admissible kernel, then the right-hand side is at most
\(\varepsilon\) above the unrestricted Kantorovich optimum.
\end{proposition}\end{mybox}
Proposition~\ref{thm:correction_ot_equivalence}, \textit{proof} in Appendix \ref{app_proof}, justifies the correction map as a
principled search over couplings rather than an auxiliary imputation heuristic.
If the class of admissible kernels is unrestricted, minimizing the expected
mechanism cost over \(T\) is exactly equivalent to minimizing the Kantorovich
cost between \(P_{\tilde{\boldsymbol X}_{\zeta}}\) and \(Q_\Theta\). In practice,
Cheesefill uses a neural family \(T_\psi\), which amortizes this coupling search:
each pseudo complete trajectory is mapped directly to a distribution over
refined trajectories. This restriction yields the upper bound in
Eq.~\eqref{eq:parametric_upper_bound}. The bound becomes tight when the neural
family is expressive enough to realize an optimal admissible coupling, and it is
approximately tight when it contains an \(\varepsilon\)-optimal admissible
coupling. Therefore, the stochastic correction map provides a principled way to
move beyond fixed imputation while still optimizing a relaxation of
the original OT objective. Cheesefill optimizes the following equation:
\begin{equation}
\min_{\Theta,\mathcal{G},\psi}\;
\mathbb{E}_{\tilde{\boldsymbol X}\sim P_{\tilde{\boldsymbol X}_{\zeta}}}
\mathbb{E}_{\boldsymbol Y\sim T_\psi(\cdot\mid\tilde{\boldsymbol X})}
\bigl[
c_{\mathcal G}(\tilde{\boldsymbol X},\boldsymbol Y)
\bigr],\quad \text{subject to } \mathbf{G}_0 \in \mathrm{DAGs}
\quad\text{and}\quad
\psi\in\Psi_{\mathrm{adm}} .
\label{eq:G_cheesefill_cost}
\end{equation}
\section{\underline{\texttt{C}}ausal 
\underline{\texttt{he}}t\underline{\texttt{e}}roscedastic 
\underline{\texttt{s}}tructure l\underline{\texttt{e}}arning 
\underline{\texttt{f}}rom \underline{\texttt{i}}ncomp\underline{\texttt{l}}ete 
tempora\underline{\texttt{l}} data}
\label{cheesefillsection}
We now instantiate the correction-map formulation with learnable components.
Cheesefill uses Bayesian structure learning to average over plausible temporal
graphs, a Temporal Graph Neural Network to parametrize the conditional mean
mechanisms, and conditional normalizing flows to model heteroscedastic residual
densities
\(\rho_\Theta^{i}(x\,;\,\mathbf p^{<t},\mathbf p^{t},\mathcal G)\).
We then describe how the instantaneous acyclicity constraint and the correction
map admissibility condition are enforced in the final optimization objective.

\paragraph{Bayesian structure learning.} Optimizing Eq.~\eqref{eq:G_cheesefill_cost} directly yields a single graph estimate. This is brittle under missingness and heteroscedastic noise: the missing entries create uncertainty about which causal relations are supported, while the noise level varies across time and samples. Hence, optimizing Eq.~\eqref{eq:G_cheesefill_cost} over a deterministic single graph can therefore concentrate too early on edges that explain imputation artifacts or local fluctuations, instead of stable causal mechanisms. To reduce this sensitivity, Cheesefill follows the Bayesian structure learning strategy of \citep{geffner2022deep, gong2022rhino, rahmaniflow}. Instead of committing to one graph during training, we maintain a variational posterior \(q_\phi(\mathcal G)\) over temporal graphs and optimize the mechanism cost averaged over graphs sampled from this posterior. Edges supported by consistent mechanism-level evidence tend to receive stable posterior mass, whereas unstable edges are averaged over competing graph samples. Since the exact graph posterior is intractable, \(q_\phi(\mathcal G)\) is parameterized as a factorized distribution over edge variables and estimated using a single Monte Carlo sample with the Gumbel Softmax reparameterization \citep{jang2016categorical}. Additional details are given in
Appendix~\ref{app:var_in_details}.
\begin{equation}
\min_{\Theta,\psi,\phi}\;
\mathbb{E}_{\tilde{\boldsymbol X}\sim P_{\tilde{\boldsymbol X}_{\zeta}}}
\mathbb{E}_{\boldsymbol Y\sim T_\psi(\cdot\mid\tilde{\boldsymbol X})}
\bigl[
\bar c_{\phi}(\tilde{\boldsymbol X},\boldsymbol Y)
\bigr]
-
\beta \mathcal{H}(q_{\phi}), \text{subject to } \mathbf{G}_0\in\mathrm{DAGs}
\text{ and }
\psi\in\Psi_{\mathrm{adm}},
\label{eq:cheesefill_cost}
\end{equation}
where
$
\bar c_{\phi}(\tilde{\boldsymbol X},\boldsymbol Y)
:=
\mathbb{E}_{\mathcal G\sim q_\phi(\mathcal G)}
\bigl[
c_{\mathcal G}(\tilde{\boldsymbol X},\boldsymbol Y)
\bigr],
$
$
\mathcal{H}(q_{\phi})
=
-\mathbb{E}_{q_\phi}[\log q_\phi],
$and \(\beta>0\) controls the entropy regularization strength. The entropy term prevents the variational posterior from collapsing too early onto a single low cost graph, which would remove the benefit of graph averaging during the initial stages of training. As evidence accumulates, \(q_\phi\) can concentrate on graphs whose mechanisms consistently explain the corrected trajectories. The averaged cost \(\bar c_\phi\) evaluates how well a pair \((\tilde{\boldsymbol X},\boldsymbol Y)\) is explained across graphs considered plausible by the posterior, rather than under a single graph estimate. Finally, \textcolor{RoyalBlue}{the equivalence and parametric upper bound results of Proposition~\ref{thm:correction_ot_equivalence} remain valid when \(c_{\mathcal G}\) is replaced by \(\bar c_\phi\),} the formal statement is given in Appendix~\ref{app:avg_costtheo_detail}.

\paragraph{Mechanism-level density parametrisation.}
Optimizing Eq.~\eqref{eq:cheesefill_cost} requires evaluating the local
mechanism density \(\rho_\Theta^i\) for arbitrary pairs
\((\tilde{\boldsymbol X},\boldsymbol Y)\), where \(\tilde{\boldsymbol X}\)
provides the child values being scored and \(\boldsymbol Y\) provides the
candidate parent values. For a graph \(\mathcal G\), let
\(\mathbf p_i^{<t}(\boldsymbol Y,\mathcal G)\) and
\(\mathbf p_i^{t}(\boldsymbol Y,\mathcal G)\) denote, respectively, the lagged
and instantaneous parents of \(x_t^i\) read from \(\boldsymbol Y\). We write the
mechanism density through a residual model:
$r_t^i
=
\tilde x_t^i
-
f_\Theta^i
\bigl(
\mathbf p_i^{<t}(\boldsymbol Y,\mathcal G),
\mathbf p_i^{t}(\boldsymbol Y,\mathcal G)
\bigr),$ so that
$\rho_\Theta^i
\bigl(
\tilde x_t^i;
\mathbf p_i^{<t},
\mathbf p_i^{t},
\mathcal G
\bigr)
=
p_\eta^i
\bigl(
r_t^i
\mid
\mathbf p_i^{<t},
\mathbf p_i^{t},
\mathcal G
\bigr),$ where $p^{i}_{\eta}$ is the density function of $g^i_{\Theta}(\mathbf{p}^{<t} , \mathbf{p}^{t})\cdot \epsilon^i$. Hence the graph-averaged mechanism cost becomes
\begin{equation}
\begin{aligned}
\bar c_{\phi}(\tilde{\boldsymbol X},\boldsymbol Y)
=
\mathbb{E}_{\mathcal G\sim q_\phi(\mathcal G)}
\left[
-\sum_{t=1}^{|\mathcal T|}
\sum_{i=1}^{d}
\log
p_\eta^i
\left(
\tilde x_t^i
-
f_\Theta^i(\mathbf p_i^{<t},\mathbf p_i^t)
\mid
\mathbf p_i^{<t},
\mathbf p_i^t,
\mathcal G
\right)
\right].
\end{aligned}
\label{eq:avg_mechanism_cost}
\end{equation}

We parameterize the mean mechanism \(f_\Theta^i\) with a temporal graph neural
network inspired by \citep{geffner2022deep, rahmaniflow}. The architecture must
respect the candidate graph: if \(y_{t-\tau}^j\) is not a parent of \(x_t^i\),
then \(f_\Theta^i\) should be insensitive to \(y_{t-\tau}^j\). We enforce this
by masking messages with the temporal adjacency matrices:
\begin{equation}
    \scalebox{1}{$f^{i}_{\Theta}\left(\mathbf{p}^{<t} , \mathbf{p}^{t}\right)=\colorboxed[brightgreen]{black}{\chi}\left(\colorboxed[brown]{black}{\sum_{\tau=0}^L \sum_{j=1}^d G_{\tau, j i}} \colorboxed[brightgreen]{black}{\vartheta}\left(y_{t-\tau}^j, \colorboxed[orrange]{black}{ e_{\tau,j}}\right), 
\colorboxed[orrange]{black}{ e_{0,i}}\right),$}
\label{neural_sem_archi}
\end{equation}
\begin{wrapfigure}{r}{0.45\textwidth}
\begin{center}
    \includegraphics[width=0.4\textwidth]{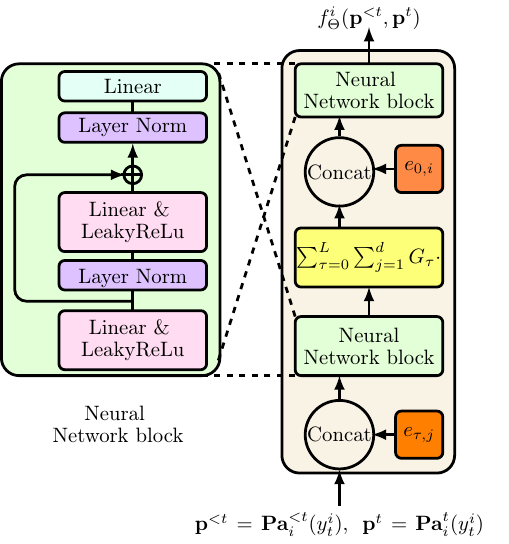}
  \end{center}
  \vspace{-10pt}
  \caption{Temporal graph neural network (TGNN) used by Cheesefill.}
  \vspace{-2pt}
  \label{archi_Cheesefill_tgnn}
\end{wrapfigure}
where $\chi$ and $\vartheta$ are neural network blocks illustrated in Figure \ref{archi_Cheesefill_tgnn} with all the other colored blocks. The trainable embeddings \(e_{\tau,j}\) encode both the variable identity
and the lag, allowing parameters to be shared across nodes and lags instead of
using a separate neural network for each mechanism. Figure~\ref{archi_Cheesefill_tgnn}
illustrates this temporal graph neural architecture. Because the adjacency
coefficient \(G_{\tau,ji}\) masks the corresponding message, absent edges do not
contribute to \(f_\Theta^i\), yielding
\(\partial f_\Theta^i / \partial y_{t-\tau}^j = 0\) whenever
\(G_{\tau,ji}=0\).

To model heteroscedastic noise, we parameterize the conditional residual density
\(p_\eta^i\) with a conditional normalizing flow. Specifically, for each
variable \(i\), we introduce an invertible map $\kappa_\Theta^i
\left(
\cdot;
\mathbf p_i^{<t},
\mathbf p_i^t,
\mathcal G
\right)
:
\mathbb R\rightarrow \mathbb R$ such that
\[
n_t^i
=
\kappa_\Theta^i
\left(
r_t^i;
\mathbf p_i^{<t},
\mathbf p_i^t,
\mathcal G
\right),
\qquad
n_t^i\sim \mathcal N(0,1).
\]
We implement \(\kappa_\Theta^i\) as a conditional spline flow
\citep{durkan2019neural}, whose spline parameters are predicted from the same
graph masked parent representation as in Eq.~\eqref{neural_sem_archi}. By the
change of variables formula,
\begin{equation}
p_\eta^i
\left(
r_t^i
\mid
\mathbf p_i^{<t},
\mathbf p_i^t,
\mathcal G
\right)
=
p_{n^i}(n_t^i)
\left|
\frac{\partial
\kappa_\Theta^i
\left(
r_t^i;
\mathbf p_i^{<t},
\mathbf p_i^t,
\mathcal G
\right)}
{\partial r_t^i}
\right|,
\label{eq_condtion_NF}
\end{equation}
where \(p_{n^i}\) is the standard normal density. This conditional flow lets the
shape and scale of the residual distribution vary with the instantaneous and
lagged parents, thereby capturing heteroscedastic noise. In the homoscedastic
case, we remove the parent conditioning from the flow and learn an unconditional
base residual distribution through an affine spline transformation of a standard
Gaussian. 
\paragraph{Correction map parametrization.}
We model the stochastic correction map \(T_\psi\) as a diagonal Gaussian kernel
over the missing entries only. Given a pseudo-complete trajectory
\(\tilde{\boldsymbol X}\), the network predicts a mean
\(\mu_\psi(\tilde{\boldsymbol X})\), and we sample
$\boldsymbol Y_{\mathrm{mis}}
\sim
\mathcal N
\bigl(
\mu_\psi(\tilde{\boldsymbol X}),
I
\bigr).$ The refined trajectory is then obtained by replacing only the missing entries $\boldsymbol Y
=
\boldsymbol M\odot \boldsymbol Y_{\mathrm{mis}}
+
(1-\boldsymbol M)\odot \tilde{\boldsymbol X}.$ Thus, \(T_\psi\) preserves the observed values exactly and only corrects the parts of the trajectory introduced by the initializer. The mean network \(\mu_\psi\) uses the same neural blocks as in Figure~\ref{archi_Cheesefill_tgnn}. We fix the covariance to \(I\) because the conditional normalizing flow already models the parent dependent residual scale and shape; letting \(T_\psi\) also learn an arbitrary variance would allow the correction map to absorb heteroscedasticity that should be explained by the SEM mechanism density.

Minimizing Eq.~\eqref{eq:cheesefill_cost} requires handling two constraints:
the acyclicity of the instantaneous graph and the admissibility of the correction
map. For acyclicity, we use $h(\boldsymbol G_0)
=
\operatorname{tr}\!\left(e^{\boldsymbol G_0\odot \boldsymbol G_0}\right)-d,$ where \(\odot\) denotes the Hadamard product, and enforce
\(h(\boldsymbol G_0)=0\) with an augmented Lagrangian penalty
\citep{zheng2018dags,pamfil2020Dynotears,rahmanicausal,rahmaniflow,geffner2022deep,gong2022rhino}.
For the admissibility condition \(P_{\tilde{\boldsymbol X}_\zeta}T_\psi=Q_\Theta\)
from Definition~\ref{def:admissible_kernel}, we use a relaxed distributional
matching penalty. Specifically, we add Maximum Mean Discrepancy regularization $\lambda_{\mathrm{map}}
\mathrm{MMD}
\bigl(
\boldsymbol Y^{(\ell)},
\boldsymbol Y^{(j)}_{\mathrm{SEM}}
\bigr),$ where \(\boldsymbol Y^{(\ell)}\) are corrected trajectories sampled from
\(T_\psi(\cdot\mid\tilde{\boldsymbol X})\), and
\(\boldsymbol Y^{(j)}_{\mathrm{SEM}}\) are trajectories generated by the current
SEM through $y^{i,(j)}_{t,\mathrm{SEM}} =f_\Theta^i(\mathbf p_i^{<t},\mathbf p_i^t)
+g_\Theta^i(\mathbf p_i^{<t},\mathbf p_i^t)\epsilon_t^{i,(j)},$ with \(\mathbf p_i^{<t}\) and \(\mathbf p_i^{t}\) denote, respectively, the lagged
and instantaneous parents of \(y^{i,(j)}_{t,\mathrm{SEM}}\) read from \(\boldsymbol Y^{(j)}\). Here \(\mathrm{MMD}\) denotes the empirical multi-scale RBF MMD and the final training objective is
\begin{equation*}
\begin{aligned}
\min_{\Theta,\psi,\phi, \zeta}\quad
&
\mathbb{E}_{\tilde{\boldsymbol X}\sim P_{\tilde{\boldsymbol X}_{\zeta}}}
\mathbb{E}_{\boldsymbol Y\sim T_\psi(\cdot\mid\tilde{\boldsymbol X})}
\left[
\bar c_\phi(\tilde{\boldsymbol X},\boldsymbol Y)
\right]
+
\mathbb{E}_{\mathcal G\sim q_\phi(\mathcal G)}
\left[
\lambda_s\|\boldsymbol G_{0:L}\|_F^2
+
\varrho h^2(\boldsymbol G_0)
+
\alpha h(\boldsymbol G_0)
\right]
\\
&+
\lambda_{\mathrm{map}}
\sum_{j,\ell = 1}^N
\mathrm{MMD}
\bigl(
\boldsymbol Y^{(\ell)},
\boldsymbol Y^{(j)}_{\mathrm{SEM}}
\bigr)-\beta \mathcal H(q_\phi).
\end{aligned}
\label{optimoverall}
\end{equation*}
\section{Experiments}\label{sec:experiments}
\subsection{Synthetic data}
\vspace{-5pt}
\textbf{Data generation. }We conduct extensive experiments to evaluate Cheesfill's performance on synthetic datasets. For ground truth graph generation, we use the Erdos–Rényi model (degree 4) for instantaneous links and the Erdos–Rényi model (degree 2) for time-lagged relationships. For data generation process, $f^i, g^i$ are chosen to be randomly initialized MLPs with one hidden layer and activation function $\{\texttt{Tanh}\}$. We evaluate the different models on MTS with missing values under heteroscedastic noise, details in Appendix \ref{synthetic_data_gen}. We consider a complex settings with multiple lags $L = 2$ or $3$. For each experiments, we  generate 50 MTS with length 200. We test different numbers of nodes $\{10, 20, 40\}$ and different missing mechanisms (MCAR, MAR, MNAR) with varying noise level $\{0.1, 0.2, 0.3, 0.4, 0.5\}$. Each experiments is repeated three times, resulting in over 108 distinct datasets. Our code is available at \href{https://github.com/arahmani1/cheesefill}{Cheesefill}.

\paragraph{Ablation studies. }
We investigate the contributions of the stochastic correction map and
the CNF through an ablation, enabling or disabling each
component. We evaluate graph recovery on a 20-node system with three
time lags, MCAR missingness (0.5 noise level), and heteroscedastic noise, reporting
instantaneous and lagged results separately.

\begin{table}[H]
    \centering
    \caption{Ablation of the stochastic correction map and CNF.
    Inst.\ and Lag denote instantaneous and lagged edges, respectively.
    F1, precision, and recall are expressed as percentages.
    The final row corresponds to full Cheesefill.}
    \label{tab:component_ablation}
    \small
    \setlength{\tabcolsep}{4pt}
    \begin{tabular}{cc*{8}{c}}
        \toprule
        & & \multicolumn{2}{c}{F1 (\%) $\uparrow$}
          & \multicolumn{2}{c}{Ratio $\downarrow$}
          & \multicolumn{2}{c}{Precision (\%) $\uparrow$}
          & \multicolumn{2}{c}{Recall (\%) $\uparrow$} \\
        \cmidrule(lr){3-4}
        \cmidrule(lr){5-6}
        \cmidrule(lr){7-8}
        \cmidrule(lr){9-10}
        Map & CNF
        & Inst. & Lag & Inst. & Lag
        & Inst. & Lag & Inst. & Lag \\
        \midrule
        Off & Off
        & 0.0 & 47.0 & 1.00 & 0.52
        & 0.0 & 32.0 & 0.0 & 89.0 \\
        Off & On
        & 0.0 & 0.0 & 1.00 & 1.00
        & 0.0 & 0.0 & 0.0 & 0.0 \\
        On & Off
        & 8.0 & 45.0 & 0.91 & 0.54
        & 5.0 & 32.0 & 17.0 & 77.0 \\
        \midrule
        \textbf{On} & \textbf{On}
        & \textbf{94.2} & \textbf{94.5}
        & \textbf{0.05} & \textbf{0.05}
        & \textbf{89.0} & \textbf{91.0}
        & \textbf{100.0} & \textbf{98.0} \\
        \bottomrule
    \end{tabular}
\end{table}

Table~\ref{tab:component_ablation} shows that the two components are
complementary in this setting. Removing both components yields zero
instantaneous F1 and a lagged F1 of $47\%$. Introducing the stochastic
correction map alone provides only limited instantaneous recovery
(F1 $=8\%$), while lagged F1 remains low at $45\%$. Conversely, enabling
the CNF without the correction map reduces F1 to zero for both edge
types. Combining both components produces a substantial improvement:
full Cheesefill achieves instantaneous and lagged F1 scores of $94.2\%$
and $94.5\%$, respectively, with precision of $89\%/91\%$ and recall of
$100\%/98\%$. These results support the joint formulation: neither component
in isolation reproduces the recovery performance obtained when the
stochastic correction map and CNF are used together.

\textbf{Benchmarks. }We benchmark our model against several baselines, including causal discovery methods designed for incomplete MTS, such as CUTS \citep{cheng2023cuts}. Cheesefill is also compared with models that assume complete MTS, including Fantom \citep{rahmaniflow}, PCMCI+ \citep{runge2020discovering}, Dynotears \citep{pamfil2020Dynotears}.
\textit{Given that these models cannot deal with incomplete MTS, we use TimeMixer imputer \citep{wang2023timemixer} to fill the Missing values and then training the aforementioned models on pseudo complete MTS}.
\begin{figure}
    \centering
    \includegraphics[width=1\linewidth]{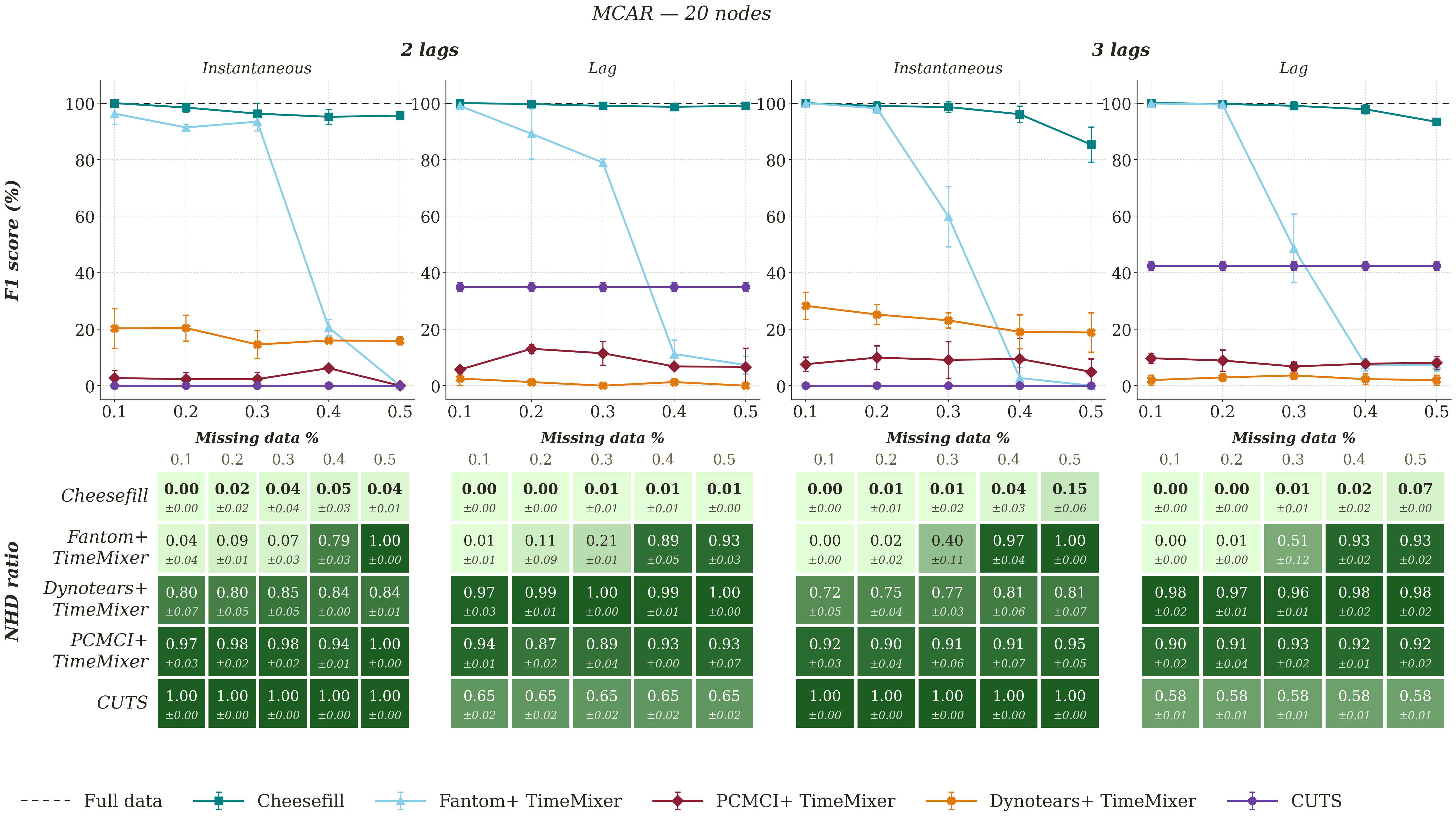}
    \caption{MCAR MTS with $20$ nodes. F1 score (top, higher is better) and NHD ratio (bottom, lower is better) for instantaneous and time lagged edges, evaluated under $2$ and $3$ maximum lags as the missing rate increases from $10\%$ to $50\%$.}
    \label{20_MCAR_synthresult}
\end{figure}

\textbf{Results and discussion.}
Figure~\ref{20_MCAR_synthresult} reports results under MCAR missingness with 20 nodes, for both 2 lag and 3 lag temporal graphs. Cheesefill stays close to the full-data reference across missing rates for both instantaneous and lagged links, and keeps the NHD ratio near zero in almost all settings. In contrast, the \textit{impute then discover} baselines degrade as the missing rate increases, showing that accurate completion alone does not necessarily preserve the causal dependencies needed for graph recovery. This effect is particularly clear for Fantom+TimeMixer: Fantom is the closest baseline to our setting because it can model instantaneous links and heteroscedastic noise, yet its performance collapses at high missingness after being fed imputed trajectories. DYNOTEARS+TimeMixer and PCMCI+TimeMixer remain far below Cheesefill throughout. CUTS does not model instantaneous links and assumes homoscedastic noise; this explains its weak performance in the instantaneous links. Overall, these results support the need for a joint correction and discovery objective rather than a pipeline that treats imputed values as fixed causal evidence.\\
Appendices~\ref{hetero} provides additional results under heteroscedastic noise, MAR and MNAR missing mechanisms, showing that Cheesefill remains robust across different data generating regimes and scales to graphs with 40 nodes. 
\subsection{Real world datasets}
\textbf{NetSim. }We evaluate Cheesefill on NetSim, a standard benchmark for temporal causal discovery \citep{gong2022rhino, cheng2023cuts}. Each MTS corresponds to the simulated BOLD signal of one subject across \(d=15\) brain regions. We use the \texttt{Sim-3.mat} setting (50 MTS, \(T=200\) timestamps, shared summary causal graph) and recover the effective connectivity from incomplete observations under varying missingness levels.\\
\begin{wrapfigure}{r}{0.4\textwidth}
  \begin{center}
    \includegraphics[width=0.4\textwidth]{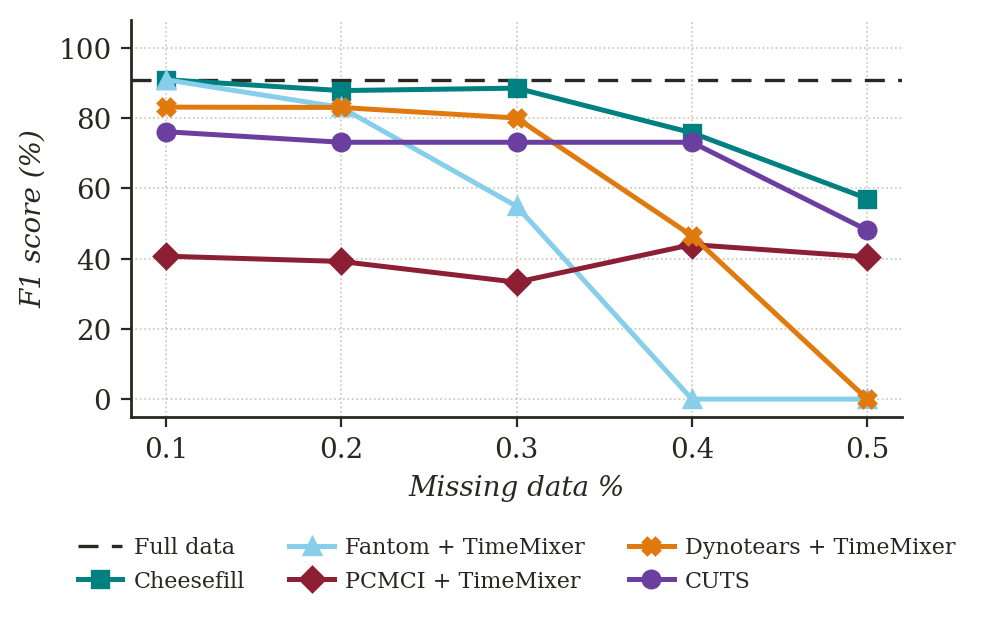}
  \end{center}
  \caption{F1 score on NetSim across missingness levels.}
  \label{nestim}
\end{wrapfigure}
Figure~\ref{nestim} shows that Cheesefill is the most robust method as missingness increases. At low missing rates several baselines remain competitive, but their performance degrades sharply beyond \(30\%\): FANTOM+TimeMixer and DYNOTEARS+TimeMixer collapse to near zero at high missing rates, while PCMCI+TimeMixer stays consistently low throughout. CUTS, designed for incomplete MTS, is more stable than the impute then discover baselines but trails Cheesefill at every level an expected gap, since CUTS models only time-lagged dependencies and assumes homoscedastic noise, an assumption known to be violated in fMRI data \citep{seiler2017multivariate}. In contrast, Cheesefill stays close to the full-data reference up to \(30\%\) missingness and retains the highest F1 score even at \(50\%\), indicating that its stochastic correction map preserves causal information that impute-then-discover pipelines lose under severe missingness.
\begin{wraptable}{r}{5.5cm}
\caption{Performance on Wind Tunnel data evaluated on summary causal graph.}
\begin{adjustbox}{center}
{\fontsize{8pt}{7pt}\selectfont 
\begin{tabular}{@{}c|cc|@{}}
           Method &F1$\uparrow$   & Ratio$\downarrow$  \\ \midrule
PCMCI+ TimeMixer        & 21.8  & 0.87            \\ 
Dynotears + TimeMixer     &   0    & 1               \\ 
Fantom +  TimeMixer    & 30.1  &   0.69              \\ 
CUTS &   19.5   &  0.89              \\ 
\rowcolor{gray!20} Cheesefill    &  \textbf{38.1} & \textbf{0,61}         \\ \bottomrule
\end{tabular}}
\end{adjustbox}
\label{table2}
\end{wraptable}

\textbf{Wind Tunnel.}
We use the wind tunnel datasets from Gamella et al.~\citep{gamella2025causal}, featuring two controllable fans pushing air through a chamber, barometers measuring air pressure at various locations, and a hatch controlling an external opening. The dataset comprises 16 variables of 10,000 samples each (see Appendix~\ref{causalchambers_app}). We compare Cheesefill to the aforementioned baselines, with results in Table~\ref{table2}. Under 50\% missingness, Cheesefill outperforms all baselines by achieving 38.1\% F1 score. This benchmark is challenging because the causal graph comes from a real physical system with coupled dynamics, indirect effects, and measurement noise, making spurious associations difficult to filter out. The gain of Cheesefill over both CUTS and \textit{impute then discover} baselines shows that its mechanism compatible correction is beneficial in a high missingness regime.
\section{Conclusion}
We introduced Cheesefill, a causal structure learning framework 
for incomplete multivariate time series that simultaneously 
handles \textit{(i)} instantaneous and time-lagged links, 
\textit{(ii)} missing observations, and \textit{(iii)} 
heteroscedastic noise. On the theoretical side, we showed that 
optimising over stochastic correction maps is equivalent to a 
conditional kernel reformulation of the OT 
problem. Extensive experiments on synthetic and 
real world data confirm consistent gains over strong baselines 
across missingness mechanisms, lags, and noise regimes.]

\paragraph{Limitations. }Cheesefill relies on the standard assumptions of SEM based temporal causal discovery, including stationarity and causal sufficiency. These assumptions may be violated in some real world systems with regime changes, hidden confounders, or unobserved external drivers. In addition, while MCAR and MAR mechanisms are theoretically well supported under standard conditions, arbitrary MNAR missingness is not identifiable without additional assumptions on the observation process. Cheesefill is therefore best viewed as a principled and empirically robust correction framework for MNAR settings, rather than a method that guarantees recovery under all possible non-ignorable missingness mechanisms. Finally, the practical objective relaxes the exact admissibility condition of the correction map using an MMD penalty and enforces instantaneous acyclicity through an augmented Lagrangian; both relaxations can lead to suboptimal solutions when optimization is difficult or when hyperparameters are poorly chosen. Despite these limitations, the empirical results show that coupling correction with mechanism level causal learning substantially improves robustness over impute then discover pipelines across missingness levels.
\newpage
\section*{Acknowledgment}
We thank Ortal Senouf for helpful feedbacks and comments.
\bibliography{biblio}

@article{cheng2023causaltime,
  title={CausalTime: Realistically Generated Time-series for Benchmarking of Causal Discovery},
  author={Cheng, Yuxiao and Wang, Ziqian and Xiao, Tingxiong and Zhong, Qin and Suo, Jinli and He, Kunlun},
  journal={arXiv preprint arXiv:2310.01753},
  year={2023}
}

@inproceedings{pamfil2020dynotears,
  title={Dynotears: Structure learning from time-series data},
  author={Pamfil, Roxana and Sriwattanaworachai, Nisara and Desai, Shaan and Pilgerstorfer, Philip and Georgatzis, Konstantinos and Beaumont, Paul and Aragam, Bryon},
  booktitle={International conference on artificial intelligence and statistics},
  pages={1595--1605},
  year={2020},
  organization={Pmlr}
}

@article{runge2019detecting,
  title={Detecting and quantifying causal associations in large nonlinear time series datasets},
  author={Runge, Jakob and Nowack, Peer and Kretschmer, Marlene and Flaxman, Seth and Sejdinovic, Dino},
  journal={Science advances},
  year={2019},
  publisher={American Association for the Advancement of Science}
}

@article{gong2022rhino,
  title={Rhino: Deep causal temporal relationship learning with history-dependent noise},
  author={Gong, Wenbo and Jennings, Joel and Zhang, Cheng and Pawlowski, Nick},
  journal={Preprint arXiv:2210.14706},
  year={2022}
}

@article{shen2020challenges,
  title={Challenges and opportunities with causal discovery algorithms: application to Alzheimer’s pathophysiology},
  author={Shen, Xinpeng and Ma, Sisi and Vemuri, Prashanthi and Simon, Gyorgy},
  journal={Scientific reports},
  year={2020},
  publisher={Nature Publishing Group UK London}
}

@article{sachs2005causal,
  title={Causal protein-signaling networks derived from multiparameter single-cell data},
  author={Sachs, Karen and Perez, Omar and Pe'er, Dana and Lauffenburger, Douglas A and Nolan, Garry P},
  journal={Science},
  year={2005},
  publisher={American Association for the Advancement of Science}
}

@article{peters2013causal,
  title={Causal inference on time series using restricted structural equation models},
  author={Peters, Jonas and Janzing, Dominik and Sch{\"o}lkopf, Bernhard},
  journal={Advances in neural information processing systems},
  year={2013}
}

@article{white2011multiple,
  title={Multiple imputation using chained equations: issues and guidance for practice},
  author={White, Ian R and Royston, Patrick and Wood, Angela M},
  journal={Statistics in medicine},
  volume={30},
  number={4},
  pages={377--399},
  year={2011},
  publisher={Wiley Online Library}
}

@article{runge2018causal,
  title={Causal network reconstruction from time series: From theoretical assumptions to practical estimation},
  author={Runge, Jakob},
  journal={Chaos: An Interdisciplinary Journal of Nonlinear Science},
  year={2018},
  publisher={AIP Publishing}
}

@inproceedings{bhattacharya2020,
  title     = {Identification In Missing Data Models Represented By Directed Acyclic Graphs},
  author    = {Bhattacharya, Rohit and Nabi, Razieh and Shpitser, Ilya and Robins, James M.},
  booktitle = {Proceedings of the 35th Conference on Uncertainty in Artificial Intelligence},
  series    = {Proceedings of Machine Learning Research},
  volume    = {115},
  pages     = {1149--1158},
  year      = {2020},
  publisher = {PMLR}
}

@inproceedings{nabi2020,
  title     = {Full Law Identification in Graphical Models of Missing Data: Completeness Results},
  author    = {Nabi, Razieh and Bhattacharya, Rohit and Shpitser, Ilya},
  booktitle = {Proceedings of the 37th International Conference on Machine Learning},
  series    = {Proceedings of Machine Learning Research},
  volume    = {119},
  pages     = {7153--7163},
  year      = {2020},
  publisher = {PMLR}
}

@inproceedings{ma2021,
  title     = {Identifiable Generative Models for Missing Not at Random Data Imputation},
  author    = {Ma, Chao and Zhang, Cheng},
  booktitle = {Advances in Neural Information Processing Systems},
  volume    = {34},
  pages     = {27645--27658},
  year      = {2021}
}

@inproceedings{rahmaniflow,
  title={Flow based approach for Dynamic Temporal Causal models with non-Gaussian or Heteroscedastic Noises},
  author={Rahmani, Abdellah and Frossard, Pascal},
  booktitle={The Thirty-ninth Annual Conference on Neural Information Processing Systems}
}

@article{cheng2023cuts,
  title={Cuts: Neural causal discovery from irregular time-series data},
  author={Cheng, Yuxiao and Yang, Runzhao and Xiao, Tingxiong and Li, Zongren and Suo, Jinli and He, Kunlun and Dai, Qionghai},
  journal={arXiv preprint arXiv:2302.07458},
  year={2023}
}

@inproceedings{rahmani2023meta,
  title={A Meta-GNN approach to personalized seizure detection and classification},
  author={Rahmani, Abdellah and Venkitaraman, Arun and Frossard, Pascal},
  booktitle={IEEE International Conference on Acoustics, Speech and Signal Processing (ICASSP)},
  year={2023}
}

@inproceedings{rahmanicausal,
  title={Causal Temporal Regime Structure Learning},
  author={Rahmani, Abdellah and Frossard, Pascal},
  booktitle={The 28th International Conference on Artificial Intelligence and Statistics},
    year={2025},
}

@inproceedings{cheng2024cuts+,
  title={Cuts+: High-dimensional causal discovery from irregular time-series},
  author={Cheng, Yuxiao and Li, Lianglong and Xiao, Tingxiong and Li, Zongren and Suo, Jinli and He, Kunlun and Dai, Qionghai},
  booktitle={Proceedings of the AAAI Conference on Artificial Intelligence},
  volume={38},
  number={10},
  pages={11525--11533},
  year={2024}
}

@article{cini2021filling,
  title={Filling the g\_ap\_s: Multivariate time series imputation by graph neural networks},
  author={Cini, Andrea and Marisca, Ivan and Alippi, Cesare},
  journal={arXiv preprint arXiv:2108.00298},
  year={2021}
}

@article{cao2018brits,
  title={Brits: Bidirectional recurrent imputation for time series},
  author={Cao, Wei and Wang, Dong and Li, Jian and Zhou, Hao and Li, Lei and Li, Yitan},
  journal={Advances in neural information processing systems},
  volume={31},
  year={2018}
}

@article{seiler2017multivariate,
  title={Multivariate heteroscedasticity models for functional brain connectivity},
  author={Seiler, Christof and Holmes, Susan},
  journal={Frontiers in neuroscience},
  year={2017},
  publisher={Frontiers Media SA}
}

@article{huizenga1995equivalent,
  title={Equivalent source estimation of scalp potential fields contaminated by heteroscedastic and correlated noise},
  author={Huizenga, Hilde M and Molenaar, Peter CM},
  journal={Brain topography},
  year={1995},
  publisher={Springer}
}

@article{hamilton1994autoregressive,
  title={Autoregressive conditional heteroskedasticity and changes in regime},
  author={Hamilton, James D and Susmel, Raul},
  journal={Journal of econometrics},
  year={1994},
  publisher={Elsevier}
}

@inproceedings{wang2023timemixer,
  title={TimeMixer: Decomposable Multiscale Mixing for Time Series Forecasting},
  author={Wang, Shiyu and Wu, Haixu and Shi, Xiaoming and Hu, Tengge and Luo, Huakun and Ma, Lintao and Zhang, James Y and Zhou, Jun},
  booktitle={International Conference on Learning Representations (ICLR)},
  year={2024}
}

@article{vo2024optimal,
  title={Optimal transport for structure learning under missing data},
  author={Vo, Vy and Zhao, He and Le, Trung and Bonilla, Edwin V and Phung, Dinh},
  journal={arXiv preprint arXiv:2402.15255},
  year={2024}
}

@article{gunther2022conditional,
  title={Conditional independence testing with heteroskedastic data and applications to causal discovery},
  author={G{\"u}nther, Wiebke and Ninad, Urmi and Wahl, Jonas and Runge, Jakob},
  journal={Advances in Neural Information Processing Systems},
  year={2022}
}

@article{durkan2019neural,
  title={Neural spline flows},
  author={Durkan, Conor and Bekasov, Artur and Murray, Iain and Papamakarios, George},
  journal={Advances in neural information processing systems},
  year={2019}
}

@inproceedings{muzellec2020missing,
  title={Missing data imputation using optimal transport},
  author={Muzellec, Boris and Josse, Julie and Boyer, Claire and Cuturi, Marco},
  booktitle={International Conference on Machine Learning},
  pages={7130--7140},
  year={2020},
  organization={PMLR}
}

@article{kantorovich2006problem,
  title={On a Problem of Monge.},
  author={Kantorovich, LK},
  journal={Journal of Mathematical Sciences},
  volume={133},
  number={4},
  year={2006}
}

@article{geffner2022deep,
  title={Deep end-to-end causal inference},
  author={Geffner, Tomas and Antoran, Javier and Foster, Adam and Gong, Wenbo and Ma, Chao and Kiciman, Emre and Sharma, Amit and Lamb, Angus and Kukla, Martin and Pawlowski, Nick and others},
  journal={arXiv preprint arXiv:2202.02195},
  year={2022}
}

@article{jang2016categorical,
  title={Categorical reparameterization with gumbel-softmax},
  author={Jang, Eric and Gu, Shixiang and Poole, Ben},
  journal={arXiv preprint arXiv:1611.01144},
  year={2016}
}

@article{zheng2018dags,
  title={Dags with no tears: Continuous optimization for structure learning},
  author={Zheng, Xun and Aragam, Bryon and Ravikumar, Pradeep K and Xing, Eric P},
  journal={Advances in neural information processing systems},
  volume={31},
  year={2018}
}

@inproceedings{runge2020discovering,
  title={Discovering contemporaneous and lagged causal relations in autocorrelated nonlinear time series datasets},
  author={Runge, Jakob},
  booktitle={Conference on Uncertainty in Artificial Intelligence},
  year={2020},
  organization={PMLR}
}

@book{little2019statistical,
  title={Statistical analysis with missing data},
  author={Little, Roderick JA and Rubin, Donald B},
  year={2019},
  publisher={John Wiley \& Sons}
}

@inproceedings{lowe2022amortized,
  title={Amortized causal discovery: Learning to infer causal graphs from time-series data},
  author={L{\"o}we, Sindy and Madras, David and Zemel, Richard and Welling, Max},
  booktitle={Conference on Causal Learning and Reasoning},
  year={2022}
}

@inproceedings{bussmann2021neural,
  title={Neural additive vector autoregression models for causal discovery in time series},
  author={Bussmann, Bart and Nys, Jannes and Latr{\'e}, Steven},
  booktitle={Discovery Science},
  year={2021},
}

@book{peters2017elements,
  title={Elements of causal inference: foundations and learning algorithms},
  author={Peters, Jonas and Janzing, Dominik and Sch{\"o}lkopf, Bernhard},
  year={2017},
  publisher={The MIT Press}
}

@book{spirtes2000causation,
  title={Causation, prediction, and search},
  author={Spirtes, Peter and Glymour, Clark N and Scheines, Richard and Heckerman, David},
  year={2000},
  publisher={MIT press}
}

@inproceedings{haufe2010sparse,
  title={Sparse causal discovery in multivariate time series},
  author={Haufe, Stefan and M{\"u}ller, Klaus-Robert and Nolte, Guido and Kr{\"a}mer, Nicole},
  booktitle={causality: objectives and assessment},
  year={2010},
  organization={PMLR}
}

@article{song2009time,
  title={Time-varying dynamic bayesian networks},
  author={Song, Le and Kolar, Mladen and Xing, Eric},
  journal={In Advances in Neural Information Processing Systems},
  year={2009}
}

@book{newman2018networks,
  title={Networks},
  author={Newman, Mark},
  year={2018},
  publisher={Oxford university press}
}

@article{hyvarinen2010estimation,
  title={Estimation of a structural vector autoregression model using non-gaussianity.},
  author={Hyv{\"a}rinen, Aapo and Zhang, Kun and Shimizu, Shohei and Hoyer, Patrik O},
  journal={Journal of Machine Learning Research},
  year={2010}
}

@article{assaad2022survey,
  title={Survey and evaluation of causal discovery methods for time series},
  author={Assaad, Charles K and Devijver, Emilie and Gaussier, Eric},
  journal={Journal of Artificial Intelligence Research},
  year={2022}
}

@inproceedings{assaad2021mixed,
  title={A mixed noise and constraint-based approach to causal inference in time series},
  author={Assaad, Karim and Devijver, Emilie and Gaussier, Eric and Ait-Bachir, Ali},
  booktitle={ECML PKDD },
  year={2021},
}

@article{entner2010causal,
  title={On causal discovery from time series data using FCI},
  author={Entner, Doris and Hoyer, Patrik O},
  journal={Probabilistic graphical models},
  year={2010}
}

@article{kingma2014adam,
  title={Adam: A method for stochastic optimization},
  author={Kingma, Diederik P and Ba, Jimmy},
  journal={arXiv preprint arXiv:1412.6980},
  year={2014}
}

@article{gamella2025causal,
  title={Causal chambers as a real-world physical testbed for AI methodology},
  author={Gamella, Juan L and Peters, Jonas and B{\"u}hlmann, Peter},
  journal={Nature Machine Intelligence},
  volume={7},
  number={1},
  pages={107--118},
  year={2025},
  publisher={Nature Publishing Group UK London}
}
\bibliographystyle{plain}

\newpage
\section*{Table of content}
\DoToC 
\clearpage
\appendix
\section{Detailed related work}
\label{app:related_work}

\paragraph{Causal structure learning from MTS.}
Causal discovery from MTS has evolved beyond the IID setting to 
account for temporal dependencies; Assaad et al. \citep{assaad2022survey} 
provide a comprehensive survey. Most methods build on Dynamic 
Bayesian Networks, which represent discrete-time dynamics as 
directed graphical models with both lagged and contemporaneous 
edges. Three families dominate the field.

\emph{Constraint-based methods} extend conditional-independence 
testing to the temporal setting. Early work focused on lagged 
relationships only \citep{haufe2010sparse, song2009time}, with 
tsFCI \citep{entner2010causal} adapting Fast Causal Inference 
\citep{spirtes2000causation} to handle latent confounders. 
PCMCI \citep{runge2019detecting} introduced a scalable two-stage 
algorithm for high-dimensional time series, initially restricted 
to lagged links and later extended to PCMCI+ 
\citep{runge2020discovering} for contemporaneous edges. NBCB 
\citep{assaad2021mixed} combines noise-based and constraint-based 
ideas to recover summary causal graphs beyond Markov equivalence, 
even in the presence of instantaneous relationships.

\emph{Functional-model approaches} exploit identifiability under 
restricted noise distributions. VARLiNGAM 
\citep{hyvarinen2010estimation} integrates non-Gaussian 
instantaneous models with vector-autoregressive components, while 
TiMINo \citep{peters2013causal} studies nonlinear instantaneous 
effects through constrained structural equation models with 
independent noise.

\emph{Score-based methods} cast structure learning as continuous 
optimisation. DYNOTEARS \citep{pamfil2020Dynotears} leverages the 
algebraic acyclicity characterisation in \citep{zheng2018dags} to 
recover both instantaneous and lagged dependencies via an 
augmented Lagrangian, without assumptions on the underlying 
topology. Rhino \citep{gong2022rhino} extends this line to 
stationary MTS with history-dependent noise, modeling the noise 
variance as a function of lagged parents.  Most recently, FANTOM 
\citep{rahmaniflow} employs conditional normalising flows to 
handle heteroscedastic noise in both stationary and non-stationary 
regimes. Despite this variety, all of these methods assume 
fully complete data. None of these methods 
addresses heteroscedastic noise jointly with instantaneous 
links and missing observations, both of which Cheesefill targets.

\paragraph{Causal discovery from incomplete MTS.}
Real-world MTS routinely contain missing values, and the dominant 
strategy in practice is the two-stage \emph{impute-then-discover} 
pipeline: complete the data with a generic imputer 
\citep{cini2021filling, cao2018brits, wang2023timemixer} and run a 
standard causal discovery method on top. This pipeline is known 
to be unreliable: the imputed distribution generally encodes a 
different set of conditional independencies than the true 
distribution, and this mismatch propagates into the recovered 
graph \citep{cheng2023cuts, cheng2024cuts+}. A small but growing 
body of work targets joint imputation and structure learning 
under missingness. CUTS \citep{cheng2023cuts} alternates between 
a delayed-supervision imputation network and a sparse Granger 
graph fitting step, and CUTS+ \citep{cheng2024cuts+} scales this 
idea to high-dimensional MTS through a message-passing GNN and a 
coarse-to-fine discovery scheme. However, both methods restrict 
themselves to purely time-lagged relationships and homoscedastic 
noise: neither can recover instantaneous links, and neither 
accounts for the noise heteroscedasticity observed in fMRI, EEG, 
and financial data.

\paragraph{Positioning of Cheesefill.}
The literature thus splits along two axes. Methods on the 
\emph{complete-data} side handle instantaneous links, 
heteroscedastic noise, or both, but require fully complete MTS 
\citep{gong2022rhino, rahmaniflow}. Methods on the 
\emph{missing data} side handle missingness, but only for 
time lagged, homoscedastic settings 
\citep{cheng2023cuts, cheng2024cuts+}. Cheesefill closes both 
gaps simultaneously: it recovers instantaneous and time lagged links from 
incomplete MTS under homoscedastic and heteroscedastic noises.
\section{Limitations}
\label{limitations}
Cheesefill relies on the standard assumptions of SEM based temporal causal discovery, including stationarity and causal sufficiency. These assumptions may be violated in some real world systems with regime changes, hidden confounders, or unobserved external drivers. In addition, while MCAR and MAR mechanisms are theoretically well supported under standard conditions, arbitrary MNAR missingness is not identifiable without additional assumptions on the observation process. Cheesefill is therefore best viewed as a principled and empirically robust correction framework for MNAR settings, rather than a method that guarantees recovery under all possible non-ignorable missingness mechanisms. Finally, the practical objective relaxes the exact admissibility condition of the correction map using an MMD penalty and enforces instantaneous acyclicity through an augmented Lagrangian; both relaxations can lead to suboptimal solutions when optimization is difficult or when hyperparameters are poorly chosen. Despite these limitations, the empirical results show that coupling correction with mechanism level causal learning substantially improves robustness over impute then discover pipelines across missingness levels.

Spurious causality is a practical risk, so we advocate using the learned graphs as decision support rather than as ground truth. Before acting on them, especially in medical or financial settings, practitioners should adopt safeguards: (i) expert vetting of proposed edges and mechanisms; (ii) robustness checks (e.g., stability under resampling or perturbations, alternative specifications); and (iii) interventional validation when feasible.
\section{Data generation and baselines}
\subsection{Data generation}
\label{synthetic_data_gen}
We employ the Erdos–Rényi (ER) \citep{newman2018networks} model with mean degrees of 1 or 2 to generate lagged graphs, and ER model with mean degrees 4 for instantaneous graphs. The maximum number of lags, $L$, is set to 2 or 3. We experiment with varying numbers of nodes $\{20,40\}$ and different missingness mechanism MCAR, MNAR and MAR, with different level $\{10\%,20\%,30\%,40\%,50\%\}$. We generate 50 different MTS of length 200 sharing the same causal graph.

In heteroscedastic settings, noise variance shifts across both variables and observations, making the underlying DAG much harder to recover from data. Given a random set if directed acyclic graphs $\mathcal{G}$, we generate observations from the SEMs in Eq \ref{eq_sem_hetero} as follows:

\begin{equation*}
x_{t}^{i,(j)} 
= f^{i}\!\bigl(\mathbf{Pa}_i^{<t},\, \mathbf{Pa}_i^{t}\bigr) 
+ \log(1+\exp (\sigma( g^{i}\!\bigl(\mathbf{Pa}_i^{<t},\, \mathbf{Pa}_i^{t}\bigr))) 
\cdot \epsilon_t^{i,(j)},
\label{eq_sem_hetero}
\end{equation*}
where $f^{i}$, $g^{i}$ are chosen to be randomly initialized MLPs with one hidden layer of size number of nodes and \texttt{tanh} activation functions. $\epsilon_t^{i}$ follows either a normal distribution $\mathcal{N}(0,1)$ or a more complex one obtained by transforming samples from a standard Gaussian with an MLP with random weights and \texttt{sin} activation function.
\subsection{Baselines}
\textbf{Dynotears \citep{pamfil2020Dynotears}. } 
Dynotears formulates causal discovery for multivariate time series through a linear vector autoregressive (VAR) model that simultaneously captures lagged \emph{and} instantaneous causal effects.  
Its key innovation is the \emph{DAGness} penalty a smooth, continuously differentiable relaxation of the acyclicity constraint optimized via an augmented Lagrangian scheme alongside a mean squared error loss.  
Dynotears assumes complete MTS with linear causal relationships. Dynotears SEM emerges as the special case of \textsc{Cheesefill} SEM obtained by using linear component functions \(f^{i}\), fixing the noise scaling to \(g^{i}=1\) and $\epsilon_t^{i} \sim \mathcal{N}(0,1)$ in Eq(\ref{eq_sem_hetero}). Moreover Cheesefill handles missingness. For comparing with this model, we use publicly available package \texttt{causalnex}\footnote{\url{https://causalnex.readthedocs.io/en/latest/}}.

\textbf{PCMCI+ \citep{runge2020discovering}. }a scalable two-stage algorithm for time series, enabling the identification of contemporaneous causal connection. As Dynotears, PCMCI+ is a special case of \textsc{Cheesefill}, obtained by using linear or non linear component functions \(f^{i}\), fixing the noise scaling to \(g^{i}=1\) and allowing $\epsilon_t^{i}$ to follow any distribution. For the comparison, we use publicly available package \texttt{Tigramite}\footnote{\url{https://jakobrunge.github.io/tigramite/}}.

\textbf{CUTS \citep{cheng2023cuts}.}
CUTS is a neural causal discovery method designed for incomplete and irregular multivariate time series. It jointly learns an imputation model and a temporal causal graph, thereby avoiding the standard impute then discover pipeline. However, CUTS focuses on time lagged Granger style dependencies and does not model instantaneous causal effects. It also assumes a homoscedastic noise structure, whereas \textsc{Cheesefill} allows the noise scale \(g^i\) in Eq.~\eqref{eq_sem_hetero} to depend on both instantaneous and lagged parents. Thus, CUTS can be viewed as addressing the missingness aspect of our problem, while \textsc{Cheesefill} further handles instantaneous links and heteroscedastic mechanisms through the stochastic correction map and conditional normalizing flows. For the comparison, we use publicly available package \texttt{CUTS}\footnote{\url{https://github.com/jarrycyx/UNN/tree/main/CUTS}}.

\textbf{Fantom \citep{rahmaniflow}.}
Fantom is a flow-based temporal causal discovery method for complete MTS with non-Gaussian and heteroscedastic noise. It models nonlinear temporal SEMs using conditional normalizing flows, allowing the residual distribution to vary with the parents and thereby capturing richer noise mechanisms than homoscedastic methods such as Dynotears and PCMCI+. Fantom is closest to \textsc{Cheesefill} on the mechanism side: in the absence of missing values, and when the pseudo complete distribution coincides with the observed complete data distribution, \textsc{Cheesefill} reduces to a related SEM-based density fitting framework with heteroscedastic mechanisms. The key difference is that Fantom assumes complete observations, whereas \textsc{Cheesefill} explicitly learns from incomplete MTS by coupling a parametric imputer with a stochastic correction map, so that imputation and graph discovery are optimized jointly rather than sequentially. For the comparison, we use publicly available code in openreview supplementary materials \texttt{Fantom}\footnote{\url{https://openreview.net/forum?id=BgwtOwFSvY}}.

\subsection{Optimization parameters}
\label{optim_param} 
The variational posterior $q_{\phi^r}(\mathcal{G}^r)$ is initialized to prefer sparse graphs (edge probability $<0.5$).  
Heteroscedastic noise is modeled with conditional normalizing flows (CNFs). Every neural block is a two-layer MLP with 32 hidden units, residual connections, and layer normalization. The neural block used for stochastic map is 3 layers MLP with 16 hidden units, residual connections, and layer normalization. We use a batch size of 1024, 2 max augmented lagrangien step and 3000 inner steps.

Gradients for discrete edges are estimated via the Gumbel–Softmax trick, using a hard forward pass and a soft backward pass with temperature~0.25.  
All spline flows employ 16 bins, and each transformation uses an embedding dimension equal to the number of nodes.

For graphs with 20 nodes we use $\rho = 0.01$ and $\alpha = 0$, the sparsity penalty is fixed at $\lambda_s = 50$, whereas for 40 nodes we set $\rho = 0.00001$ and we use a sparsity penalty of $\lambda_s = 30$.  
Models are optimized with Adam~\citep{kingma2014adam} at a learning rate of~0.005. 
\section{Real world data}
\subsection{Causal Chambers data}
\label{causalchambers_app}
\begin{figure}[H]
    \centering
    \includegraphics[width=0.9\linewidth]{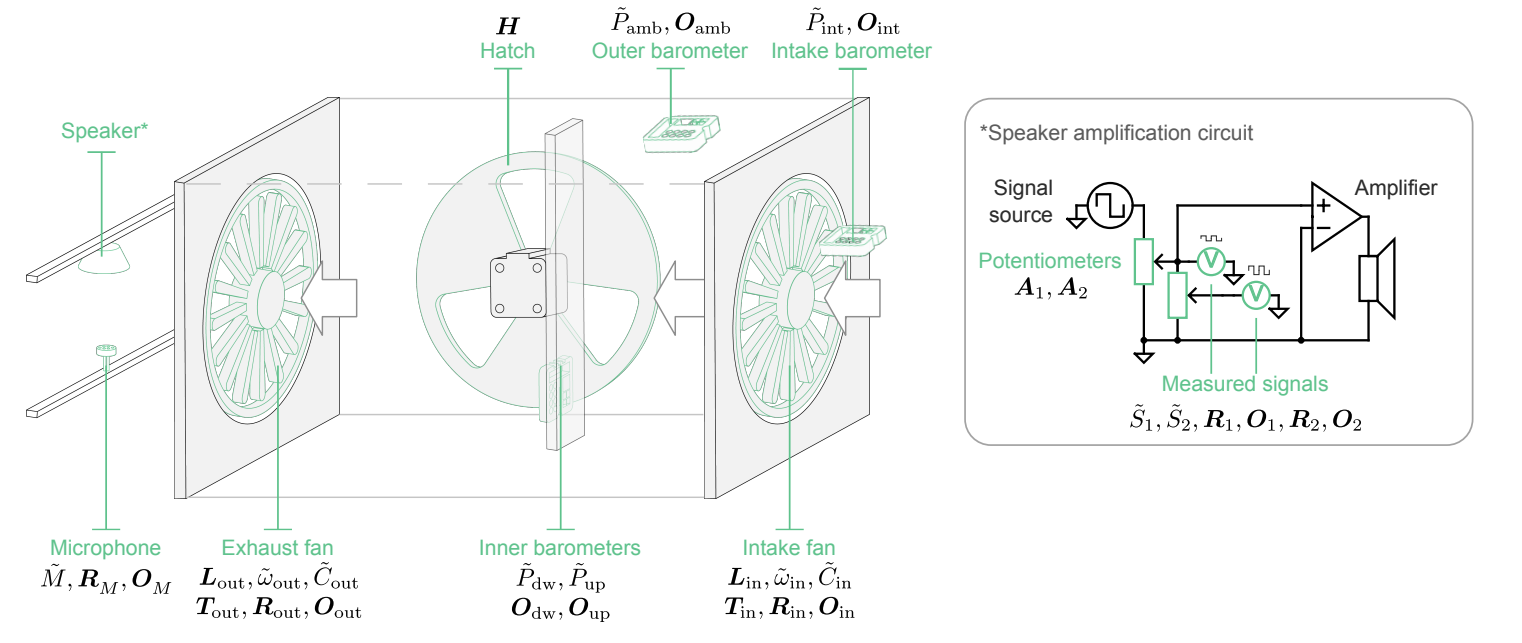}
    \caption{Figure taken from Gamella et al. \citep{gamella2025causal}. Diagrams of the wind tunel causal chamber and its main components, including the amplification circuit
that drives the speaker of the wind tunnel. The variables measured by
the chamber are displayed in black math print. Sensor measurements are denoted by a tilde. Manipulable variables, that is, actuators and sensor parameters, are shown in bold symbols. }
    \label{fig:causalchambers}
\end{figure}

We use the wind tunnel datasets from Gamella et al. \citep{gamella2025causal}, featuring two controllable fans pushing air through a chamber, barometers measuring air pressure at various locations, and a hatch controlling an external opening. The tunnel is a chamber with two controllable fans that push air through it and barometers that measure air pressure at different locations. A hatch precisely controls the area of an additional opening to the outside (see Figure \ref{fig:causalchambers}).
The dataset comprises 16 variables: controllable load of the two fans $L_{in}, L_{out}$, their measurable speed $\left(\tilde{\omega}_{\text {in }}, \bar{\omega}_{\text {out }}\right)$, the current draw by the fans $\left(\tilde{C}_{\text {in }}, \tilde{C}_{\text {out }}\right)$ ( $\tilde{C}_{\text {in }}, \tilde{C}_{\text {out }}$ ), the resulting air pressure inside the chamber ( $\tilde{P}_{\text {dw }}, \tilde{P}_{\text {up }}$ ) or at its intake ( $\tilde{P}_{\text {int }}$ ), and the hatch $H$. In the circuit that drives the speaker, we can manipulate the potentiometers ( $A_1, A_2$ ) that control the amplification, monitoring the resulting signal at different points of the circuit $\left(\tilde{S}_1, \tilde{S}_2\right)$ and through the microphone output $(\tilde{M})$. $\tilde{\tilde{P}}_{\text {amb }}$ is the ambient pressure measure by the outer barometer. 
We evaluate all the models on MTS of 10,000 samples.

For the training procedure, we set our lag to 8 time lagged and we allow the presence of instantaneous parents. Regarding the parameters, we use a sparsity coefficient equal to 500 for time lagged links and 100 for instantaneous, spline of 128 bins and MLPs of size 32.
\subsection{NetSim}
\begin{figure}[t]
    \begin{minipage}[t]{.45\textwidth}
       \begin{center}
    \includegraphics[width=0.9\textwidth]{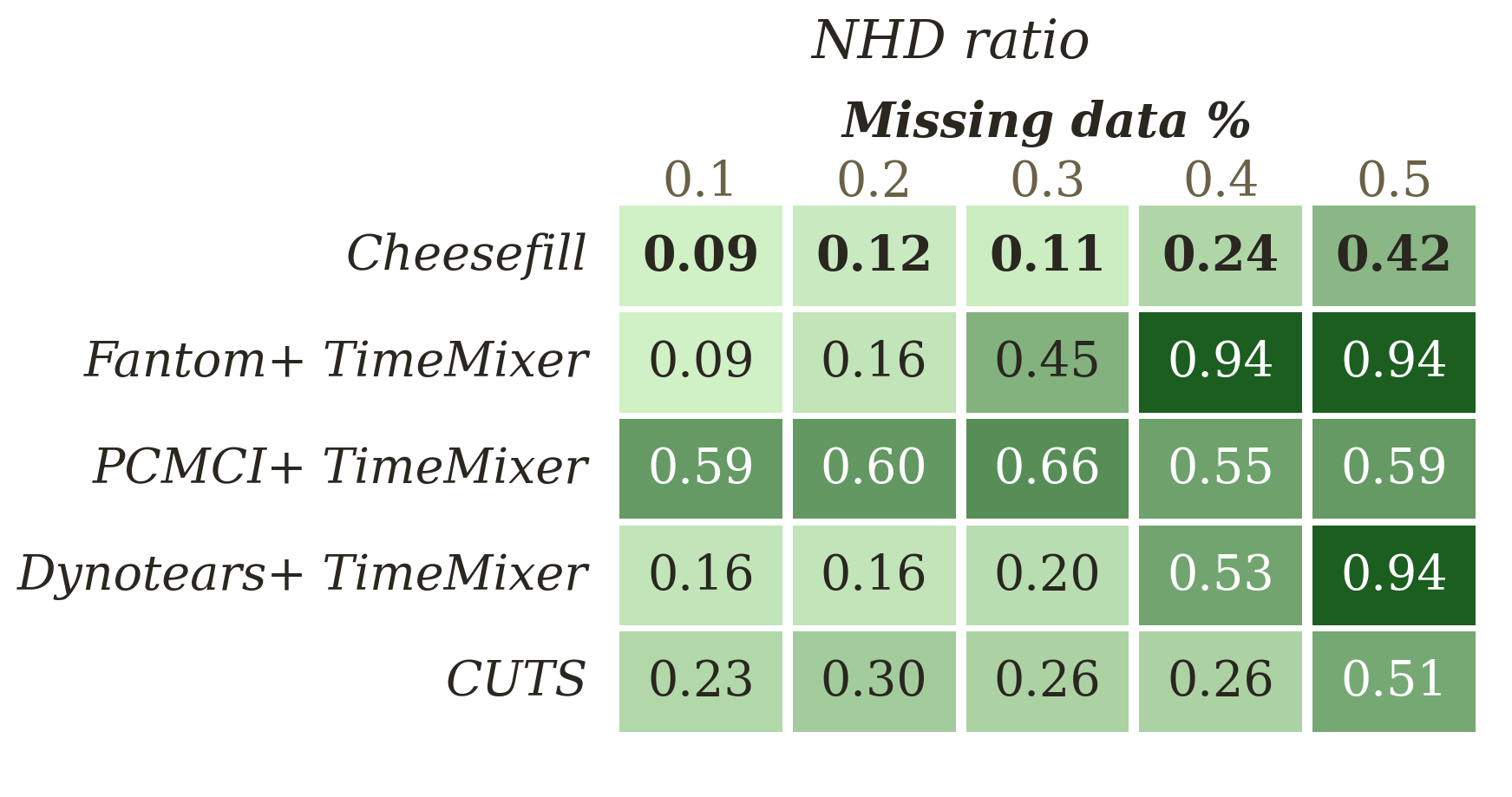}
  \end{center}
  \vspace{-10pt}
   \caption{NHD ratio on NetSim across missingness levels.}
  \vspace{-10pt}
    \end{minipage}
    \hfill
    \begin{minipage}[t]{.45\textwidth}
        \begin{center}
    \includegraphics[width=0.9\textwidth]{figures/f1_5models.png}
\end{center}
  \vspace{-15pt}
  \caption{F1 score on NetSim across missingness levels.}
  \label{pure_impure_fig}
    \end{minipage}  
\end{figure}

We evaluate Cheesefill on NetSim, a standard benchmark for temporal causal discovery \citep{gong2022rhino, cheng2023cuts}. Each MTS corresponds to the simulated BOLD signal of one subject across \(d=15\) brain regions. We use the \texttt{Sim-3.mat} setting (50 MTS, \(T=200\) timestamps, shared summary causal graph) and recover the effective connectivity from incomplete observations under varying missingness levels using MCAR mechanism.\\
Figure~\ref{nestim} shows that Cheesefill is the most robust method as missingness increases. At low missing rates several baselines remain competitive, but their performance degrades sharply beyond \(30\%\): FANTOM+TimeMixer and DYNOTEARS+TimeMixer collapse to near zero at high missing rates, while PCMCI+TimeMixer stays consistently low throughout. CUTS, designed for incomplete MTS, is more stable than the impute then discover baselines but trails Cheesefill at every level an expected gap, since CUTS models only time-lagged dependencies and assumes homoscedastic noise, an assumption known to be violated in fMRI data \citep{seiler2017multivariate}. In contrast, Cheesefill stays close to the full-data reference up to \(30\%\) missingness and retains the highest F1 score even at \(50\%\), indicating that its stochastic correction map preserves causal information that impute-then-discover pipelines lose under severe missingness. NHD ratio metrics highlight the same trends where Cheesefill outperforms all the baseline and prove that Cheesefill learns accurate summary causal graph. 
\section{Variational distribution Details}
\label{app:var_in_details}
We provide the detailed formulations for $q_{\phi}(\mathcal{G})$. In order to model the temporal adjacency matrices $\boldsymbol{G}_{\tau}$ where $\tau \in [|1:L|]$, we use two learnable matrices $\boldsymbol{U}_{\tau}, \boldsymbol{Q}_{\tau} \in \mathbf{R}^{d\times d}$ such that:
\begin{equation}
p_{k, i j}=\frac{\exp \left(u_{k, i j}\right)}{\exp \left(u_{k, i j}\right)+\exp \left(q_{k, i j}\right)}
\end{equation}
For instantaneous graphs $\boldsymbol{G}_0$, we used the same trick as in \cite{gong2022rhino, rahmaniflow}, in which we employ three lower triangular learnable matrices $\boldsymbol{U}_{0}, \boldsymbol{Q}_{0}, \boldsymbol{E}_{0} \in \mathbf{R}^{d\times d}$ to characterise three scenarios: (1) $i \rightarrow j$; (2) $j \rightarrow i$; (3) no edge between them. For node $i>j$:
\begin{equation}
\begin{aligned}
p(i \rightarrow j) & =\frac{\exp \left(u_{i j}\right)}{\exp \left(u_{i j}\right)+\exp \left(q_{i j}\right)+\exp \left(e_{i j}\right)} \\
p(j \rightarrow i) & =\frac{\exp \left(q_{i j}\right)}{\exp \left(u_{i j}\right)+\exp \left(q_{i j}\right)+\exp \left(e_{i j}\right)} \\
p(\text { no edge }) & =\frac{\exp \left(e_{i j}\right)}{\exp \left(u_{i j}\right)+\exp \left(q_{i j}\right)+\exp \left(e_{i j}\right)} .
\end{aligned}
\end{equation}
With this formulation, the instantaneous adjacency matrix is free of self-loops, eliminating any length-1 cycles.
\section{Additional Experiments}
\label{hetero}
\subsection{Additional results on synthetic data}
\subsubsection{MCAR results}
\begin{figure}[h]
    \centering
    \includegraphics[width=1\linewidth]{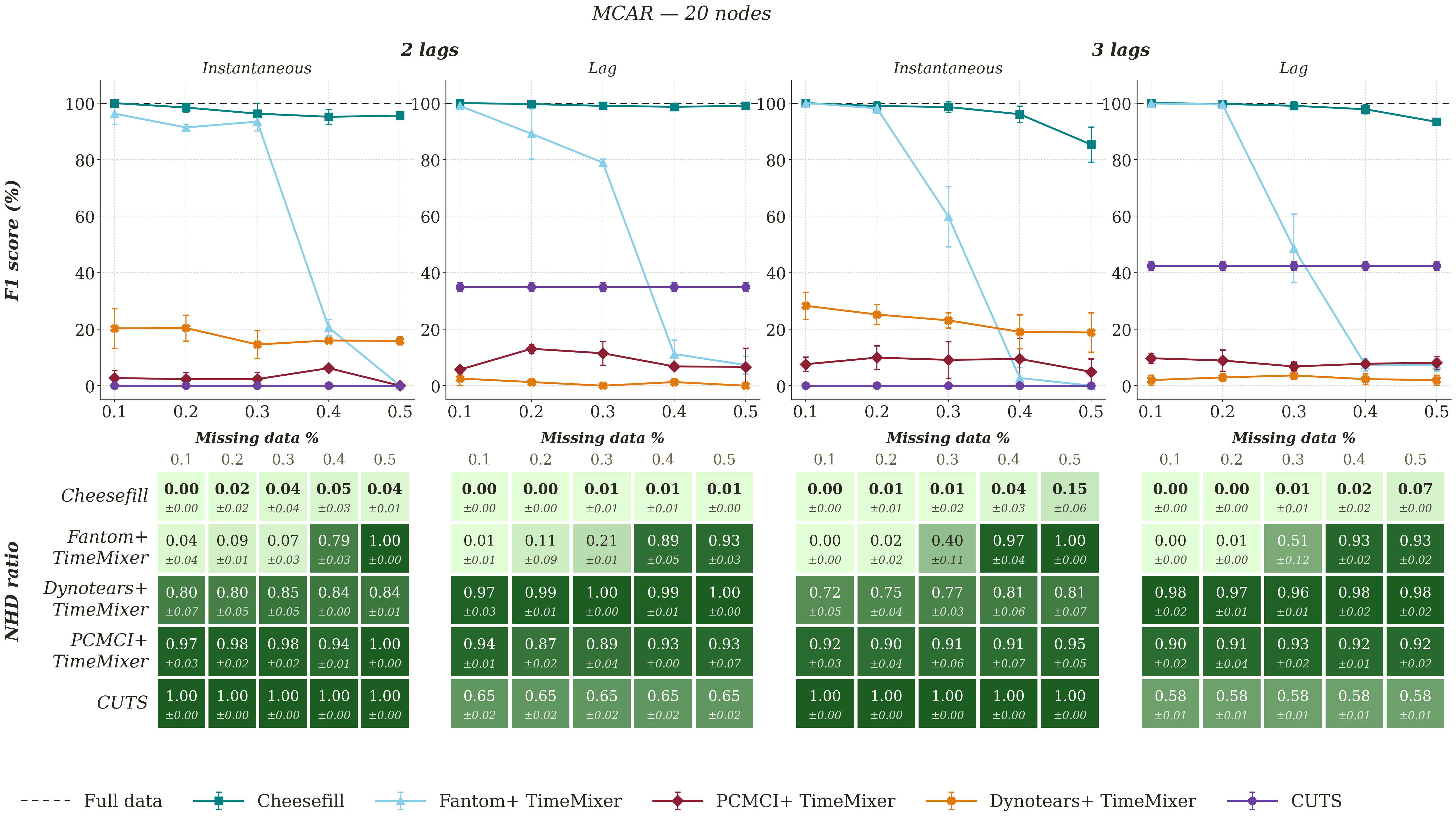}
    \caption{MCAR MTS with $20$ nodes. F1 score (top, higher is better) and NHD ratio (bottom, lower is better) for instantaneous and time lagged edges, evaluated under $2$ and $3$ maximum lags as the missing rate increases from $10\%$ to $50\%$.}
       \label{20_MCAR_synthresult_app}
\end{figure}
Figure~\ref{20_MCAR_synthresult_app} reports results under MCAR missingness with 20 nodes, for both 2 lag and 3 lag temporal graphs. Cheesefill stays close to the full-data reference across missing rates for both instantaneous and lagged links, and keeps the NHD ratio near zero in almost all settings. In contrast, the \textit{impute then discover} baselines degrade as the missing rate increases, showing that accurate completion alone does not necessarily preserve the causal dependencies needed for graph recovery. This effect is particularly clear for Fantom+TimeMixer: Fantom is the closest baseline to our setting because it can model instantaneous links and heteroscedastic noise, yet its performance collapses at high missingness after being fed imputed trajectories. DYNOTEARS+TimeMixer and PCMCI+TimeMixer remain far below Cheesefill throughout. Overall, these results support the need for a joint correction-and-discovery objective rather than a pipeline that treats imputed values as fixed causal evidence.
\begin{figure}[H]
    \centering
    \includegraphics[width=1\linewidth]{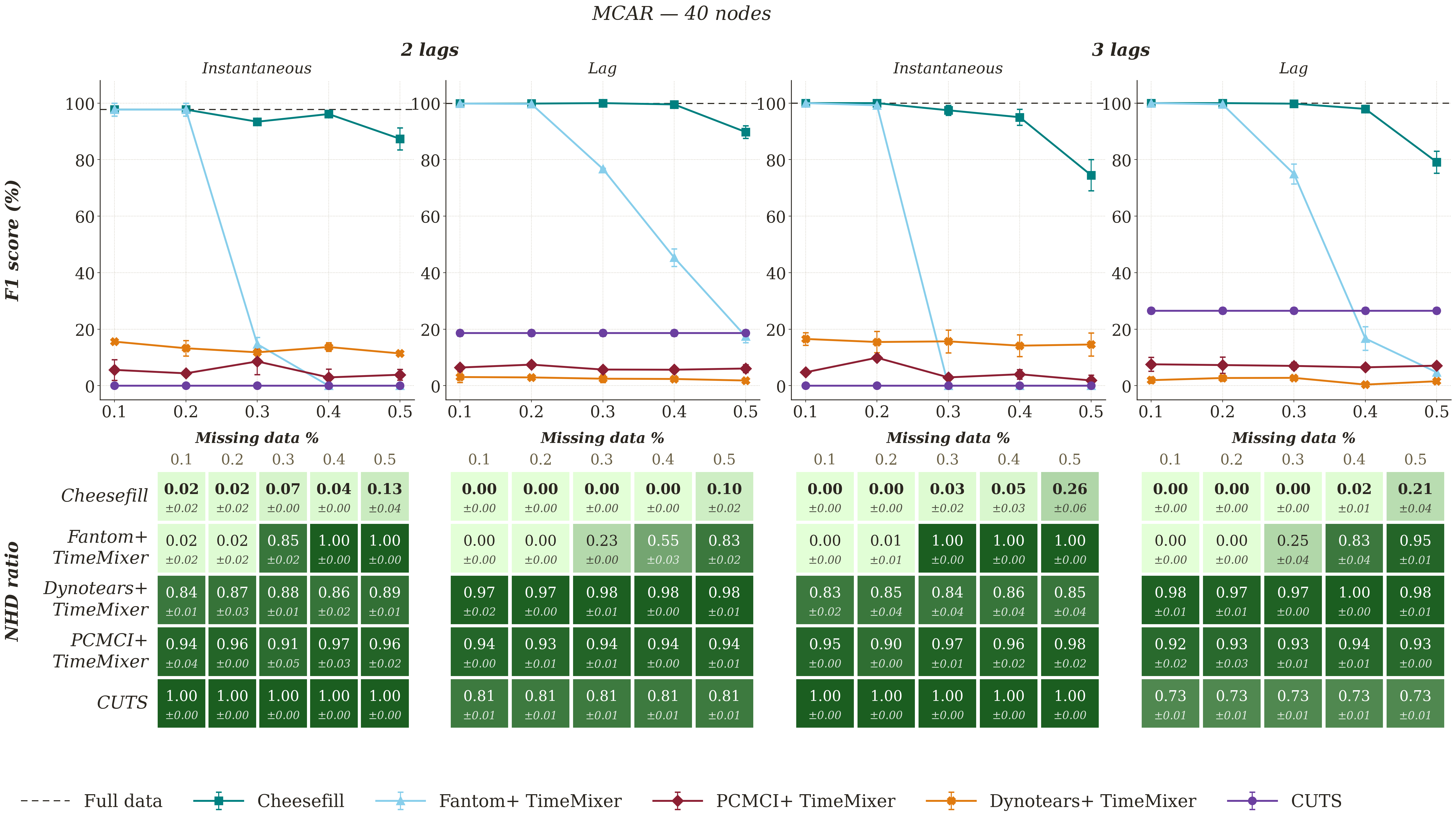}
    \caption{MCAR MTS with $40$ nodes. F1 score (top, higher is better) and NHD ratio (bottom, lower is better) for instantaneous and time lagged edges, evaluated under $2$ and $3$ maximum lags as the missing rate increases from $10\%$ to $50\%$.}
    \label{40_MCAR_synthresult}
\end{figure}
\paragraph{MCAR, 40 nodes.}
Figure~\ref{40_MCAR_synthresult} shows the corresponding 40 node MCAR results. Although MCAR is less biased than MAR, the high dimensional setting remains difficult because missing entries reduce the effective information available for recovering both instantaneous and lagged links. Cheesefill again achieves the strongest overall performance, maintaining near perfect F1 scores for most missingness levels and keeping the NHD ratio close to zero, especially for 2 lag graphs and moderate missing rates. Performance decreases only in the hardest 3 lag setting at 50\% missingness, where the graph is denser and fewer complete temporal contexts are available; nevertheless, Cheesefill still substantially outperforms all baselines. The impute then discover methods degrade rapidly as missingness increases, with Fantom+TimeMixer collapsing despite its ability to model heteroscedastic noise on complete data. These results support the central claim that causal recovery from incomplete MTS benefits from jointly correcting pseudo complete trajectories and learning SEM compatible mechanisms, rather than treating imputation as a preprocessing step.
\subsubsection{MAR results}
We set 10\% of the variables to be fully observed.
\begin{figure}[h]
    \centering
    \includegraphics[width=1\linewidth]{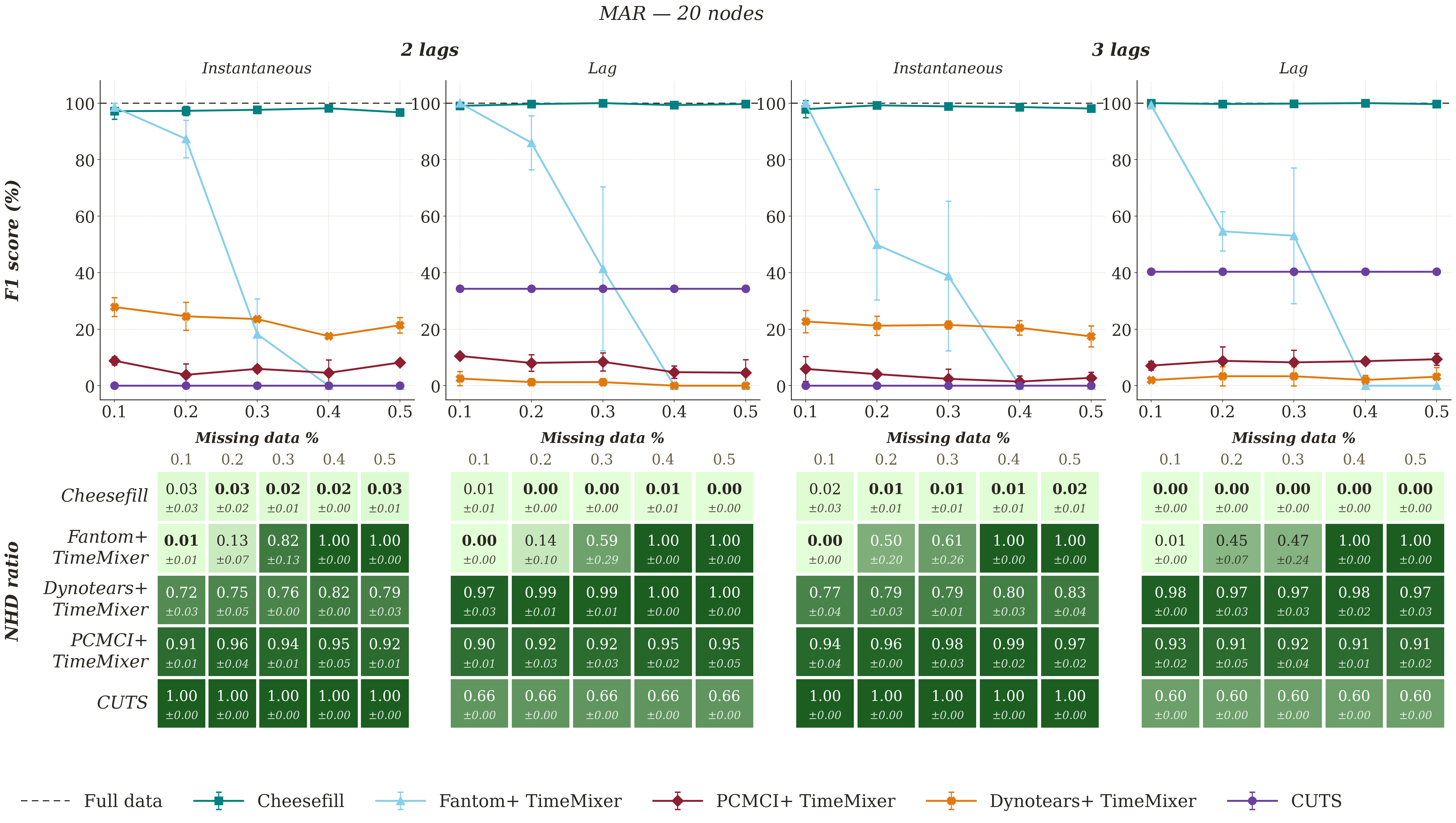}
    \caption{MAR MTS with $20$ nodes. F1 score (top, higher is better) and NHD ratio (bottom, lower is better) for instantaneous and time lagged edges, evaluated under $2$ and $3$ maximum lags as the missing rate increases from $10\%$ to $50\%$.}
    \label{20_MAR_synthresult_app}
\end{figure}
\paragraph{MAR, 20 nodes.}
Figure~\ref{20_MAR_synthresult_app} reports the MAR setting with 20 nodes. Cheesefill remains consistently close to the full-data reference across both 2 lag and 3 lag graphs, for instantaneous as well as lagged edges. Its NHD ratio stays near zero in almost all settings, showing that the recovered graph remains structurally close to the ground truth even as the missing rate increases. In contrast, Fantom+TimeMixer is competitive only at low missingness, then degrades sharply as the missing rate grows, especially for instantaneous links. This highlights the limitation of applying a strong complete-data method after fixed imputation: the imputer can introduce dependencies that are then interpreted as causal structure. Dynotears+TimeMixer and PCMCI+TimeMixer remain substantially below Cheesefill throughout, confirming that impute-then-discover pipelines are not robust under MAR missingness. CUTS is more stable on lagged edges than the impute then discover baselines, but it does not model instantaneous links and assumes homoscedastic mechanisms; this explains its weak performance in the instantaneous panels and its gap to Cheesefill in this heteroscedastic setting. Overall, these results show that jointly correcting pseudo-complete trajectories and learning SEM-compatible mechanisms is essential for reliable causal recovery under biased missingness.
\begin{figure}[h]
    \centering
    \includegraphics[width=1\linewidth]{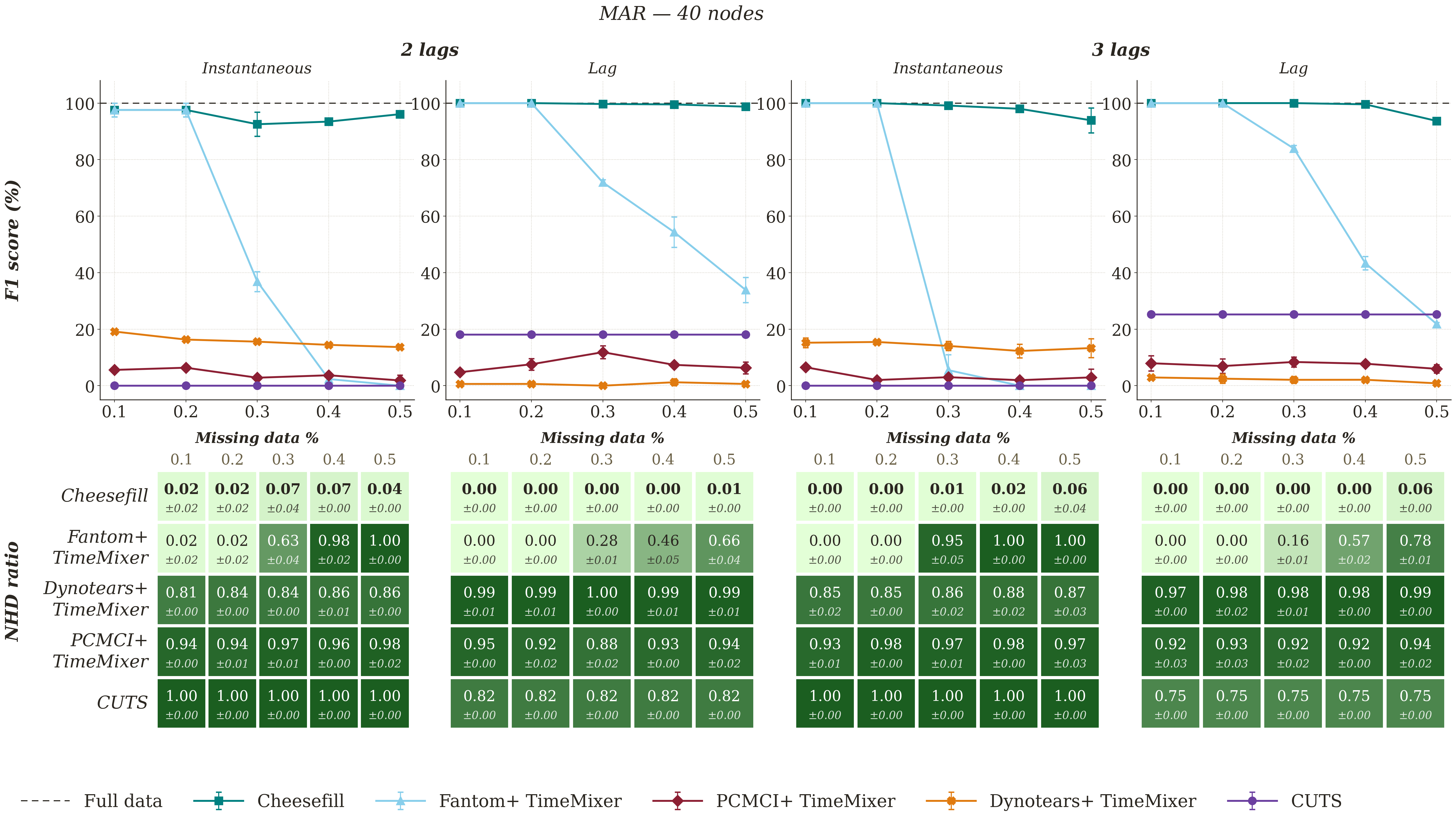}
    \caption{MAR MTS with $40$ nodes. F1 score (top, higher is better) and NHD ratio (bottom, lower is better) for instantaneous and time lagged edges, evaluated under $2$ and $3$ maximum lags as the missing rate increases from $10\%$ to $50\%$.}
    \label{40_MAR_synthresult}
\end{figure}
\paragraph{MAR, 40 nodes.}
Figure~\ref{40_MAR_synthresult} reports the 40 node MAR setting, which is substantially more challenging than MCAR because the missingness pattern depends on observed values and can therefore introduce systematic bias. Cheesefill remains close to the full-data reference across both 2 lag and 3 lag graphs, for both instantaneous and lagged edges. Its NHD ratio stays near zero in almost all settings, indicating that the recovered graph differs only marginally from the full-data reference even as the missing rate increases. In contrast, Fantom+TimeMixer performs well at low missingness but degrades sharply once the missing rate increases, especially for instantaneous edges, showing that a strong complete-data causal discovery method can still fail when it is applied after fixed imputation. Dynotears+TimeMixer, PCMCI+TimeMixer, and CUTS remain far below Cheesefill throughout, which confirms that MAR missingness requires more than either impute then discover pipelines or incomplete MTS methods restricted to lagged homoscedastic dependencies.
\section{MNAR Experiments}
\label{app_mnarresults}
MNAR is the most challenging missingness regime because the probability that an entry is missing may depend on the unobserved value itself. In general, this makes the complete data distribution non-identifiable without additional assumptions or knowledge of the missingness mechanism. In our experiments, we therefore do not treat MNAR as a theoretically identifiable setting; instead, we use a small observed subset to model the missingness process and test whether Cheesefill remains empirically robust under non-ignorable corruption. Specifically, we set 10\% of the variables to be fully observed.
\begin{figure}[h]
    \centering
    \includegraphics[width=1\linewidth]{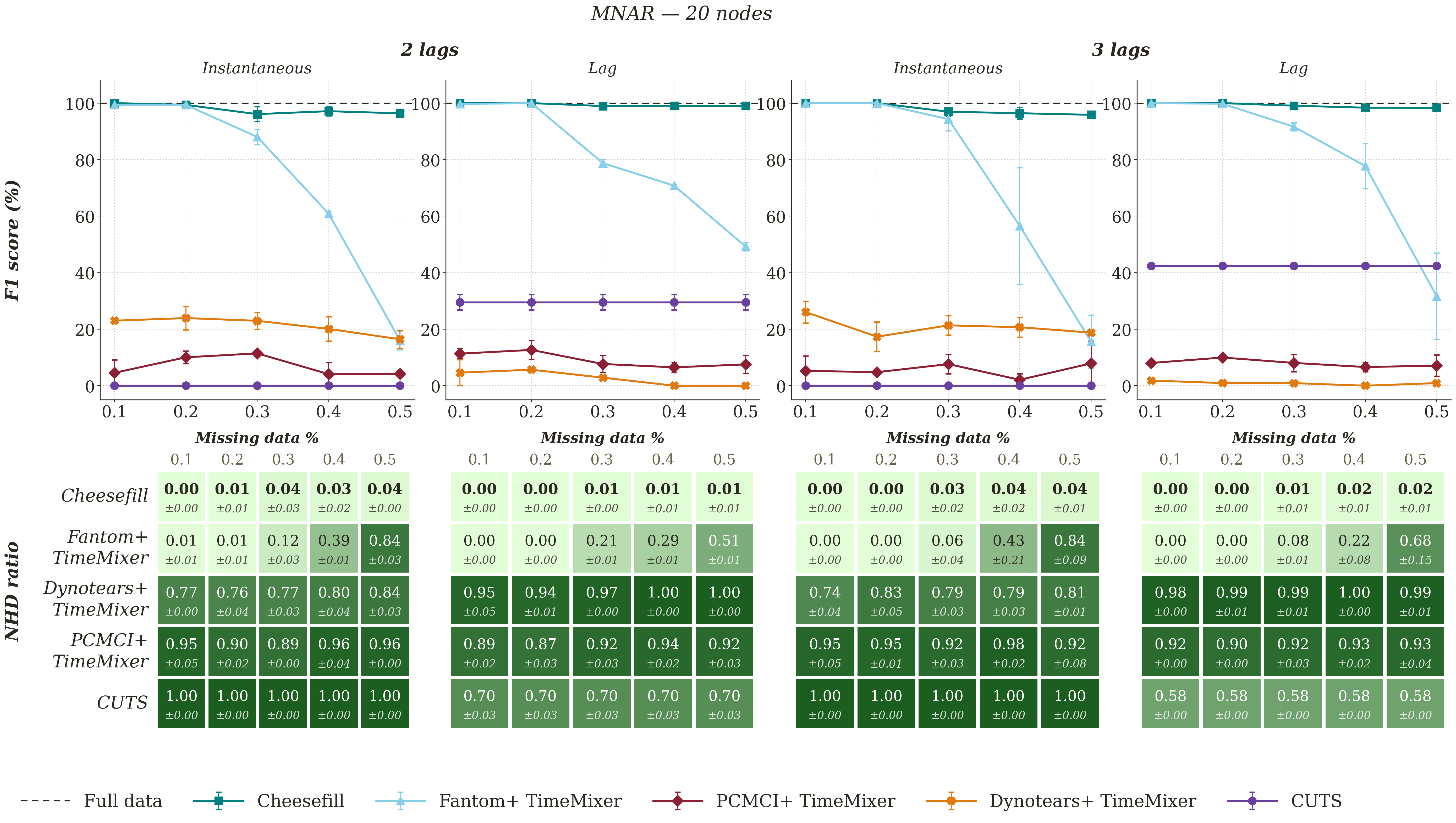}
    \caption{MNAR MTS with $20$ nodes. F1 score (top, higher is better) and NHD ratio (bottom, lower is better) for instantaneous and time lagged edges, evaluated under $2$ and $3$ maximum lags as the missing rate increases from $10\%$ to $50\%$.}
    \label{20_MNAR_synthresult}
\end{figure}
Figure~\ref{20_MNAR_synthresult} shows that Cheesefill remains consistently close to the full data reference across both 2 lag and 3 lag graphs. Its F1 score stays high for instantaneous and lagged edges even as the missing rate increases, and its NHD ratio remains near zero in almost all settings. This is notable because MNAR corruption can systematically bias the pseudo complete distribution, making fixed imputations especially unreliable. Fantom+TimeMixer is competitive at low missingness but degrades substantially at higher missing rates, particularly for instantaneous links, showing again that complete-data causal discovery methods are sensitive to imputation artifacts. Dynotears+TimeMixer and PCMCI+TimeMixer remain far below Cheesefill, while CUTS is stable only for lagged effects and cannot recover instantaneous links; it also does not model heteroscedastic mechanisms. Overall, these results show that Cheesefill’s mechanism-compatible correction map provides strong empirical robustness even in MNAR settings, where exact recovery cannot be expected without additional assumptions on the observation process.

\section{Additional experiments on non linear SEM with gaussian noise}
We sample causal graphs from the Erd\H{o}s--R\'enyi (ER) model \citep{newman2018networks}, with mean degree 4 for the instantaneous graph and mean degree 2 for each lagged graph, i.e., each variable has on average two instantaneous parents and two parents at every lag. The instantaneous graph is a DAG, obtained by sampling edges consistent with a random topological order, while lagged edges, including self-lags, are sampled independently. The maximum lag is $L=3$ and the number of nodes is $d\in\{20,40\}$. 

\paragraph{Homoscedastic case.}
Here the noise variance is the same for every variable and observation. Given the sampled graph $\mathcal{G}$, we generate observations from the additive nonlinear SEM in Eq.~\ref{eq_sem_homo}:
\begin{equation}
x_t^{i,(j)} = \sum_{k=0}^{L}\ \sum_{m\in\mathbf{Pa}_i^{t-k}} w_{mi}^{(k)}\tanh\!\bigl(x_{t-k}^{m,(j)}\bigr) + \epsilon_t^{i,(j)}, \qquad \epsilon_t^{i,(j)}\sim\mathcal{N}(0,1),
\label{eq_sem_homo}
\end{equation}
where $\mathbf{Pa}_i^{t-k}$ is the set of parents of $x_t^{i}$ at lag $k$, so that $\mathbf{Pa}_i^{t}$ ($k=0$) are the instantaneous parents, evaluated in topological order, and $\mathbf{Pa}_i^{<t}=\bigcup_{k=1}^{L}\mathbf{Pa}_i^{t-k}$ are the lagged ones. We added also a new baseline for this setting Rhino \citep{gong2022rhino}.
\subsection{MCAR and MAR results}
\begin{figure}[h]
    \centering
    \includegraphics[width=1\linewidth]{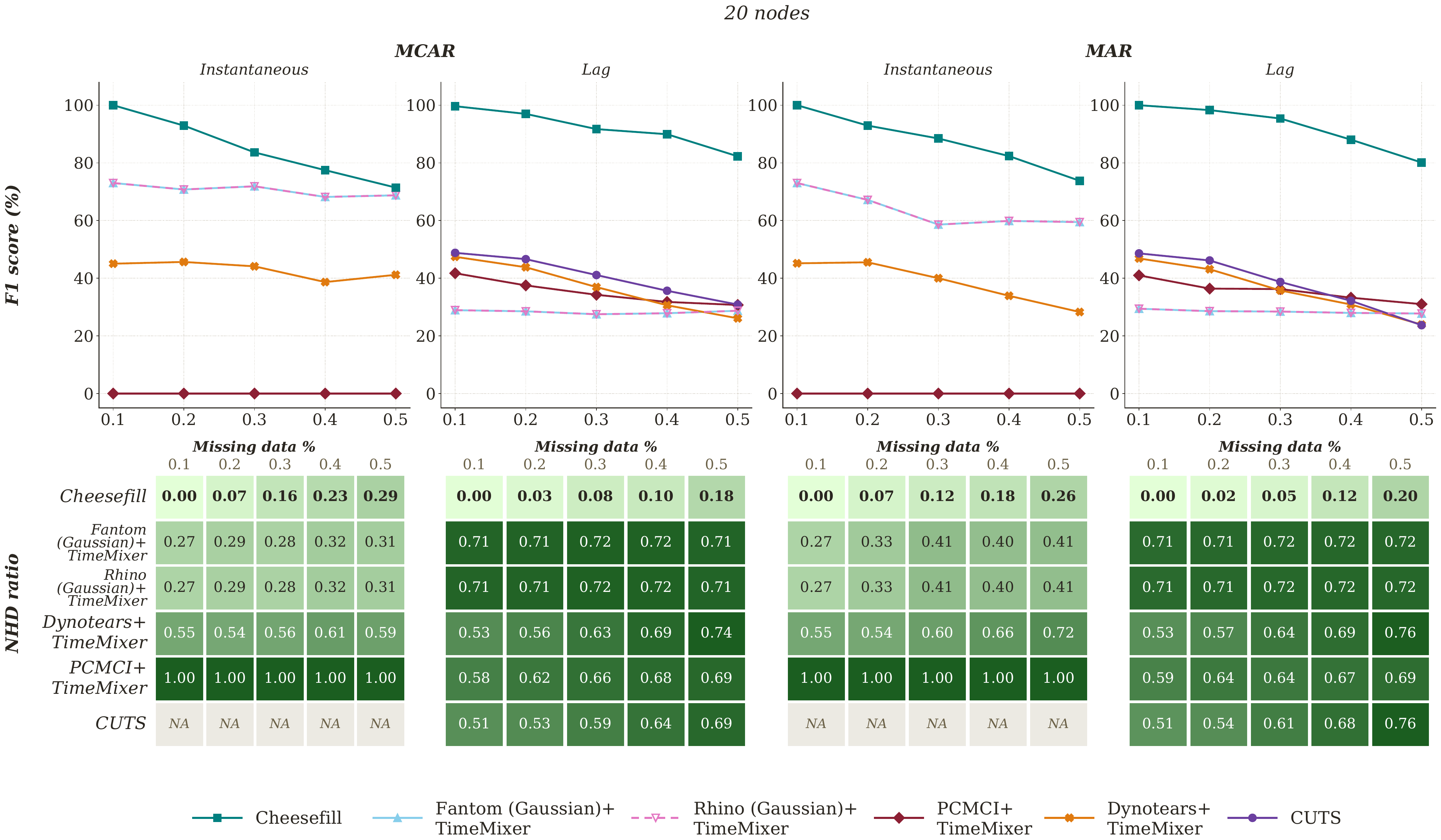}
    \caption{MCAR and MAR MTS with $20$ nodes. F1 score (top, higher is better) and NHD ratio (bottom, lower is better) for instantaneous and time lagged edges, evaluated under $3$ maximum lags as the missing rate increases from $10\%$ to $50\%$.}
    \label{20_MCAR_MAR_synthresult_app}
\end{figure}

\begin{figure}[h]
    \centering
    \includegraphics[width=0.6\linewidth]{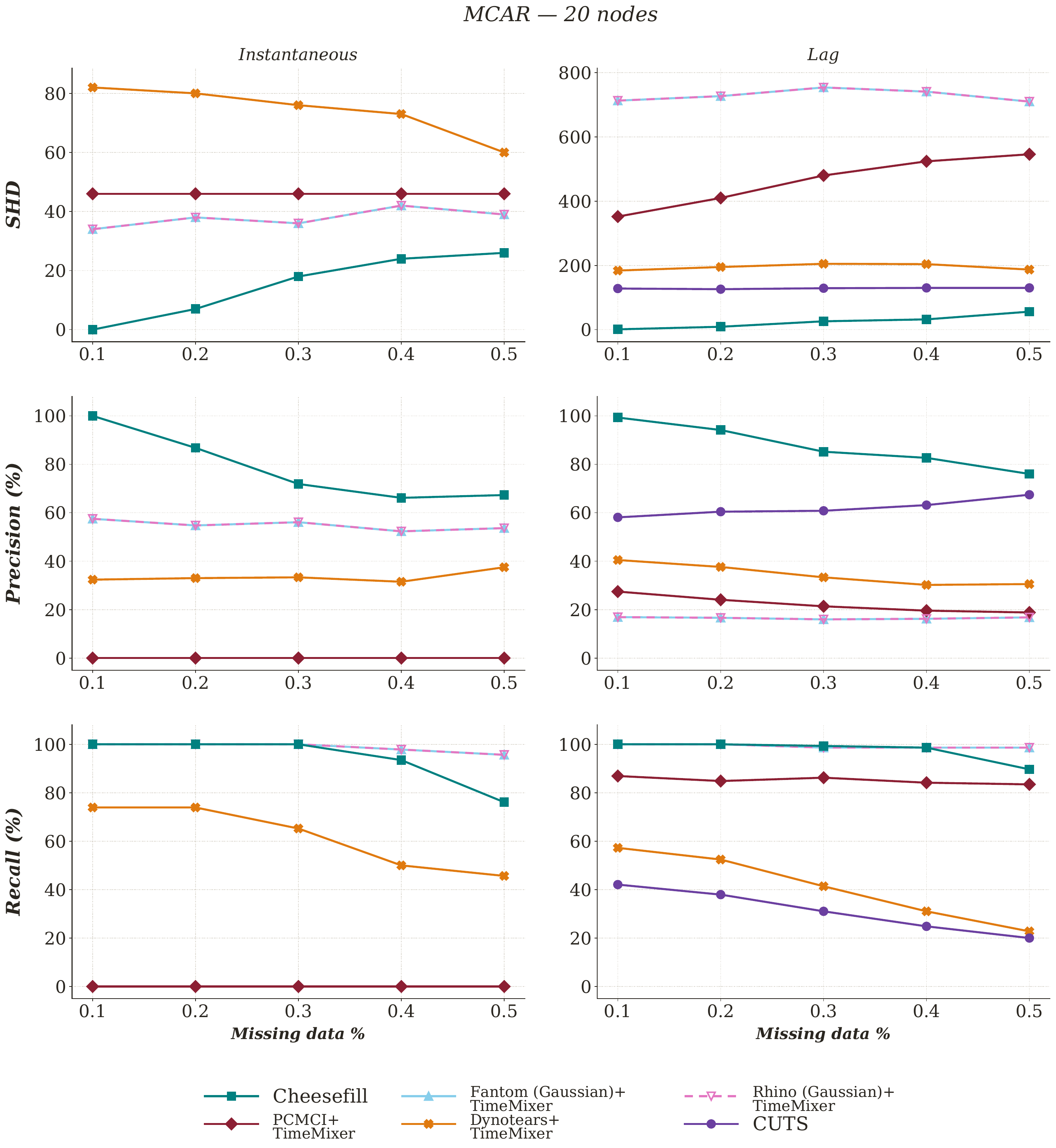}
    \caption{MCAR MTS with $20$ nodes. SHD (top, lower is better), Precision (middle, higher is better) and Recall (bottom, higher is better) for instantaneous and time lagged edges, evaluated under $3$ maximum lags as the missing rate increases from $10\%$ to $50\%$.}
    \label{20_MCAR_synthresult_app_recall}
\end{figure}
\begin{figure}[H]
    \centering
    \includegraphics[width=0.6\linewidth]{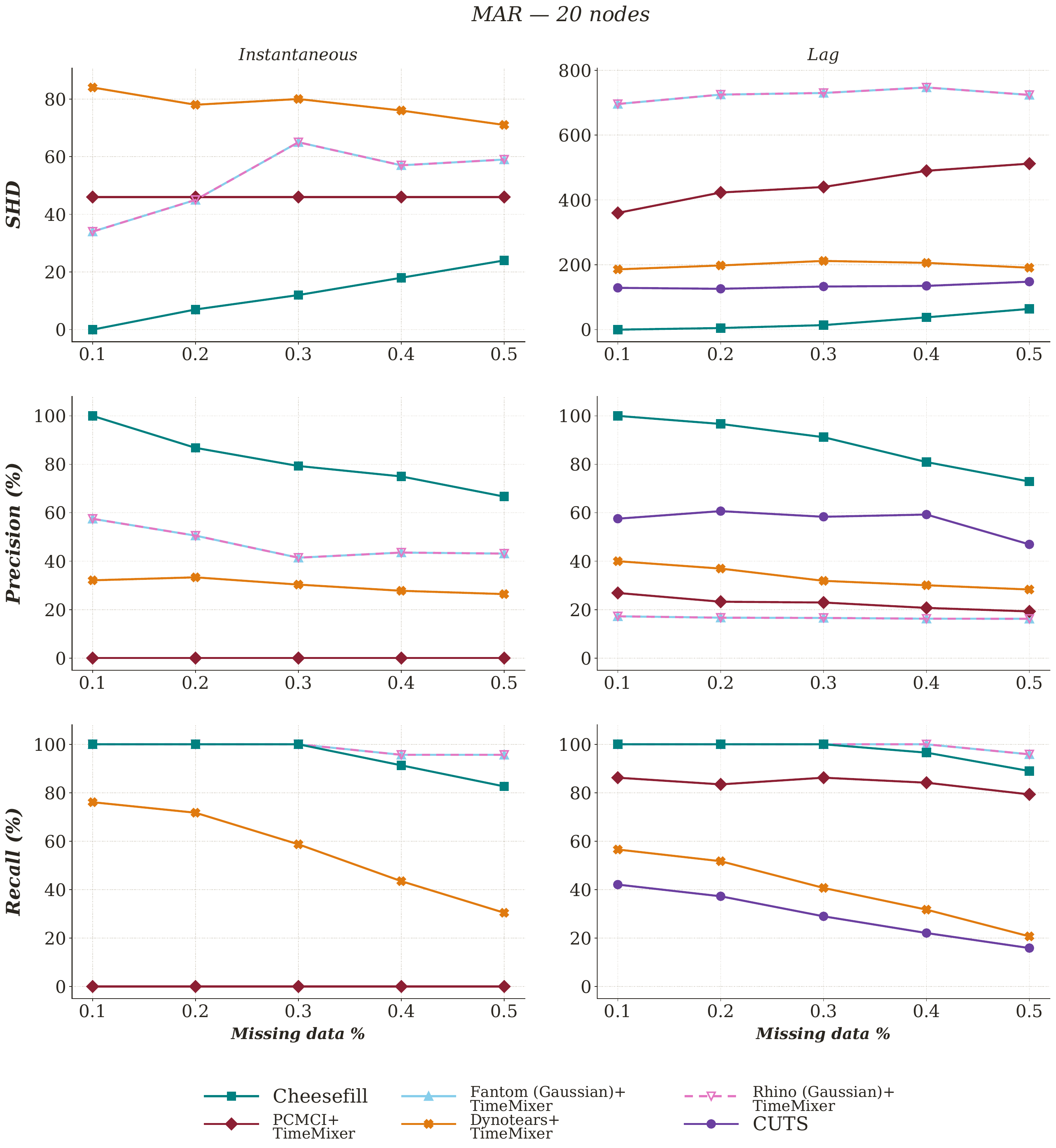}
    \caption{MAR MTS with $20$ nodes. SHD (top, lower is better), Precision (middle, higher is better) and Recall (bottom, higher is better) for instantaneous and time lagged edges, evaluated under $3$ maximum lags as the missing rate increases from $10\%$ to $50\%$.}
    \label{20_MAR_synthresult_app_recall}
\end{figure}

\begin{figure}[H]
    \centering
    \includegraphics[width=1\linewidth]{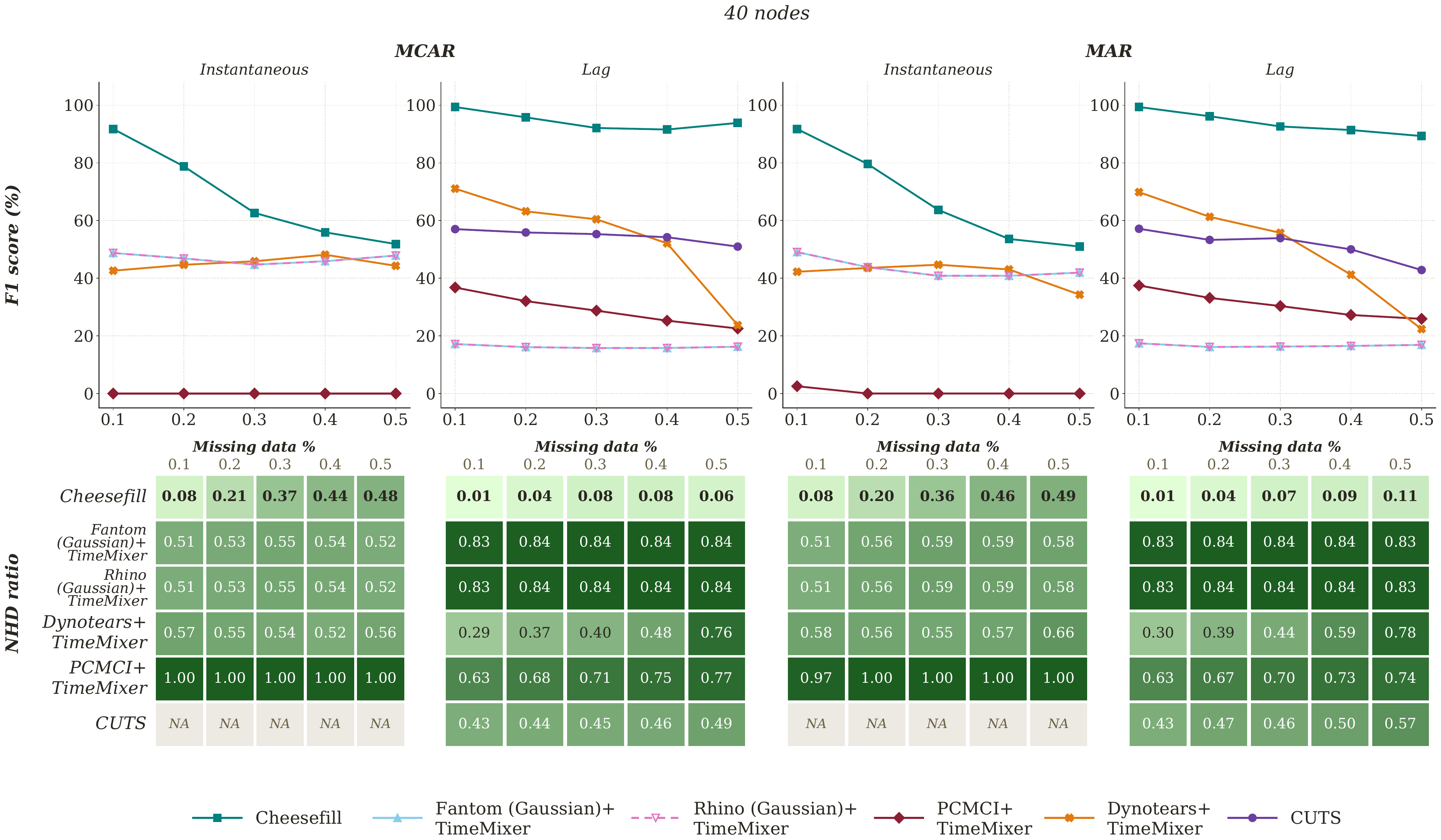}
    \caption{MCAR and MAR MTS with $40$ nodes. F1 score (top, higher is better) and NHD ratio (bottom, lower is better) for instantaneous and time lagged edges, evaluated under $3$ maximum lags as the missing rate increases from $10\%$ to $50\%$.}
    \label{40_MCAR_MAR_synthresult_app}
\end{figure}

\begin{figure}[H]
    \centering
    \includegraphics[width=0.6\linewidth]{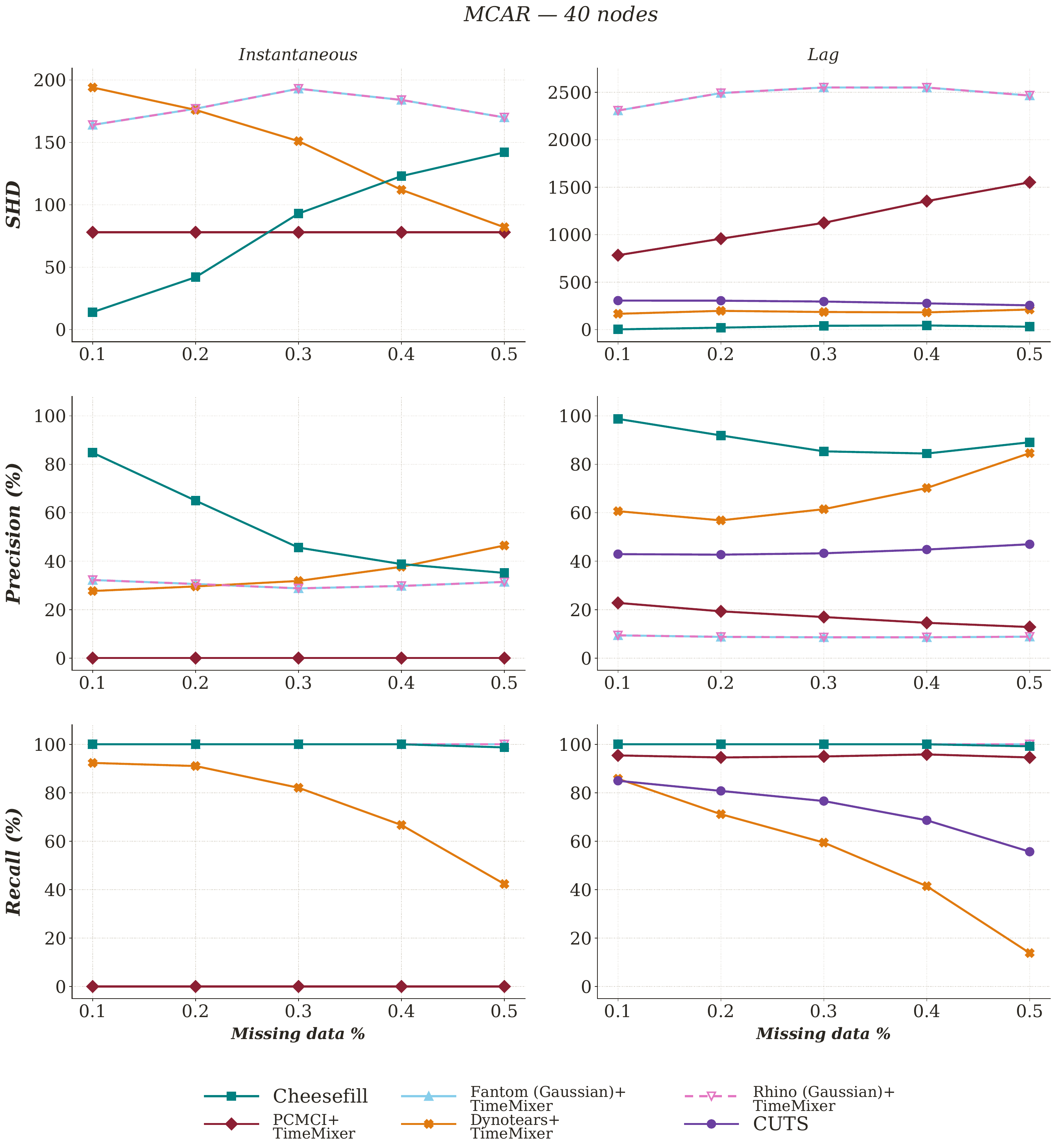}
    \caption{MCAR MTS with $40$ nodes. SHD (top, lower is better), Precision (middle, higher is better) and Recall (bottom, higher is better) for instantaneous and time lagged edges, evaluated under $3$ maximum lags as the missing rate increases from $10\%$ to $50\%$.}
    \label{40_MCAR_synthresult_app_recall}
\end{figure}
\begin{figure}[H]
    \centering
    \includegraphics[width=0.6\linewidth]{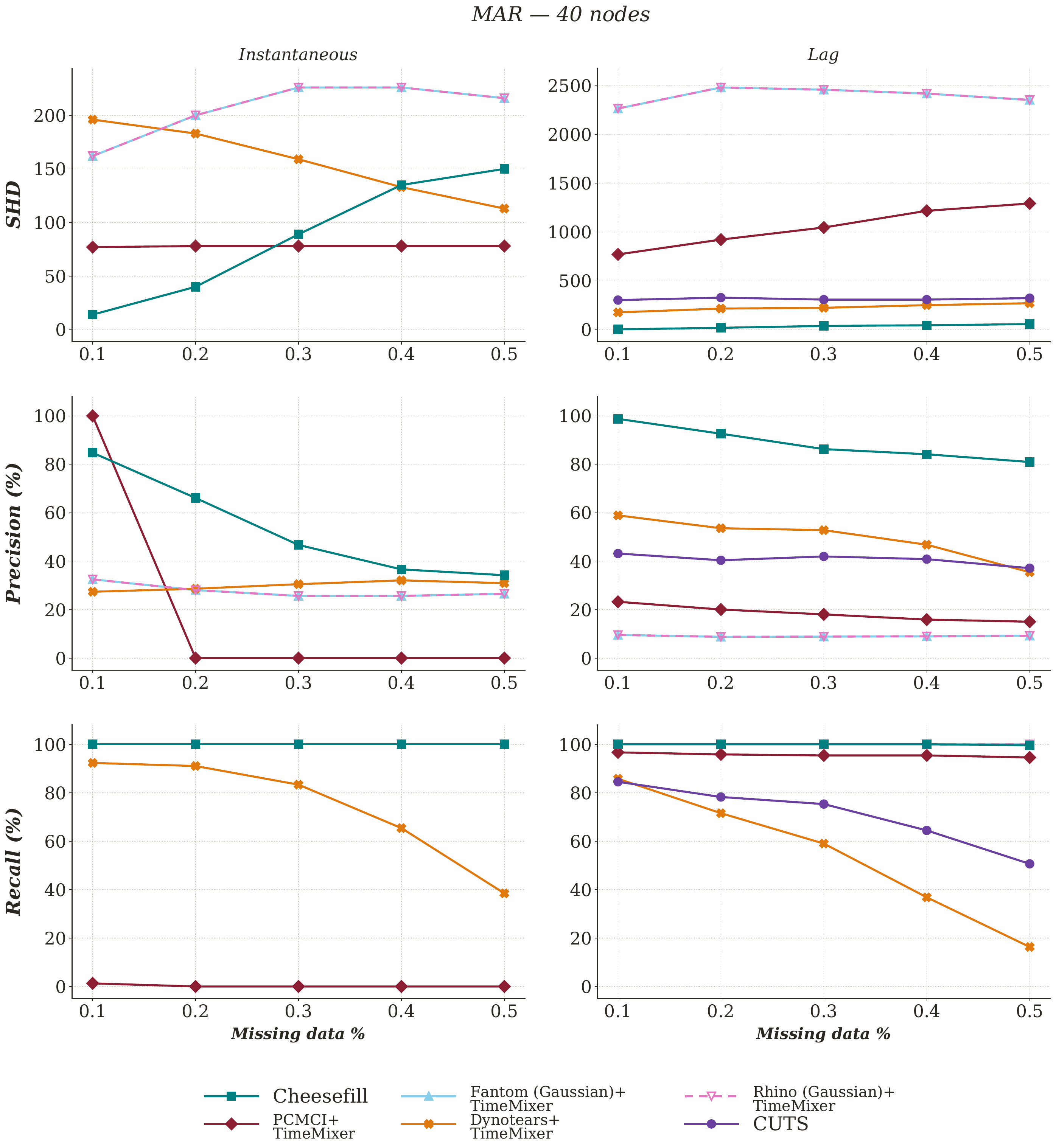}
    \caption{MAR MTS with $40$ nodes. SHD (top, lower is better), Precision (middle, higher is better) and Recall (bottom, higher is better) for instantaneous and time lagged edges, evaluated under $3$ maximum lags as the missing rate increases from $10\%$ to $50\%$.}
    \label{40_MAR_synthresult_app_recall}
\end{figure}
\subsection{MNAR results}
\begin{figure}[H]
    \centering
    \includegraphics[width=0.6\linewidth]{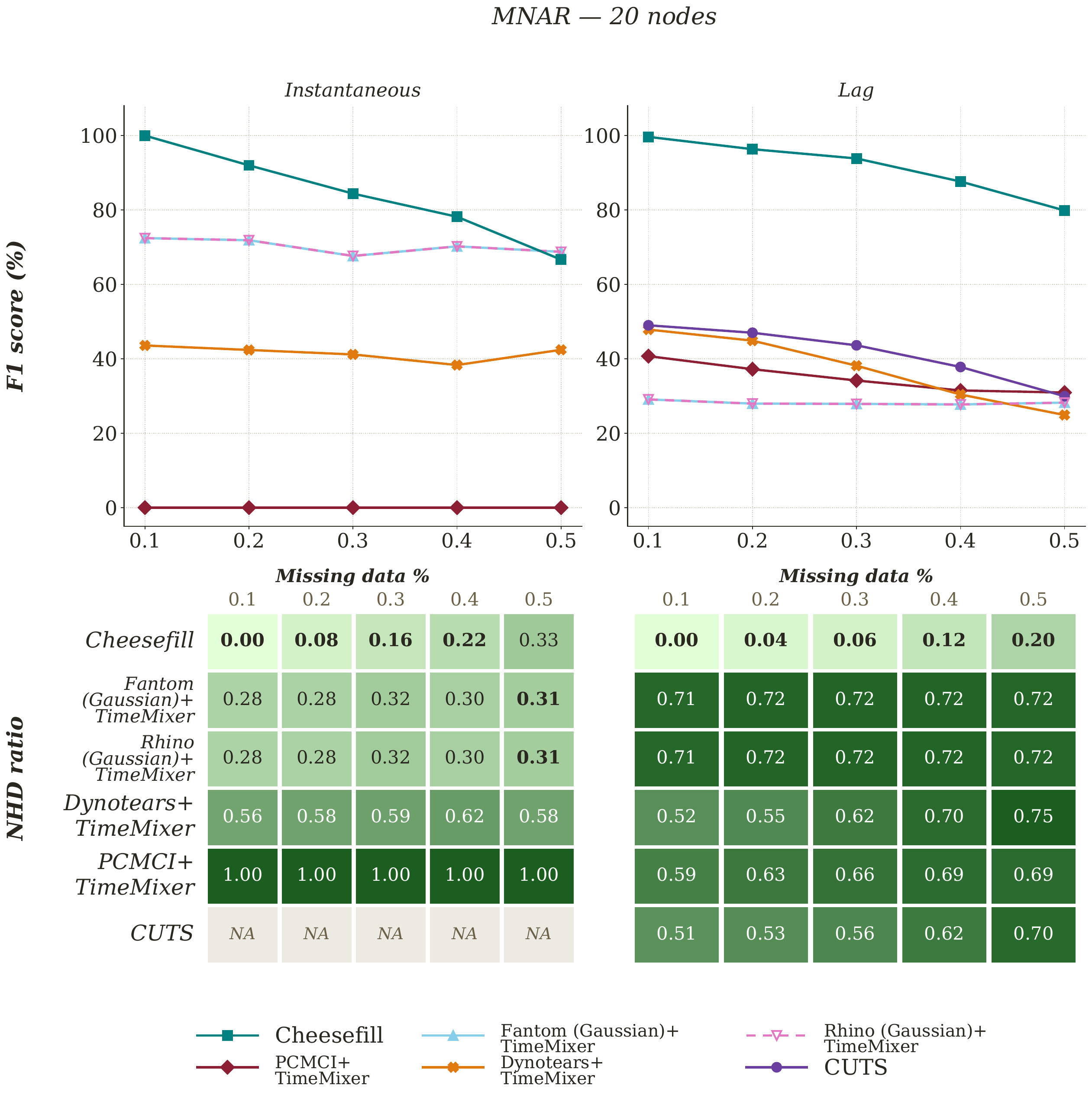}
    \caption{MNAR MTS with $20$ nodes. F1 score (top, higher is better) and NHD ratio (bottom, lower is better) for instantaneous and time lagged edges, evaluated under $3$ maximum lags as the missing rate increases from $10\%$ to $50\%$.}
    \label{20_MNAR_synthresult_app}
\end{figure}

\begin{figure}[H]
    \centering
    \includegraphics[width=0.6\linewidth]{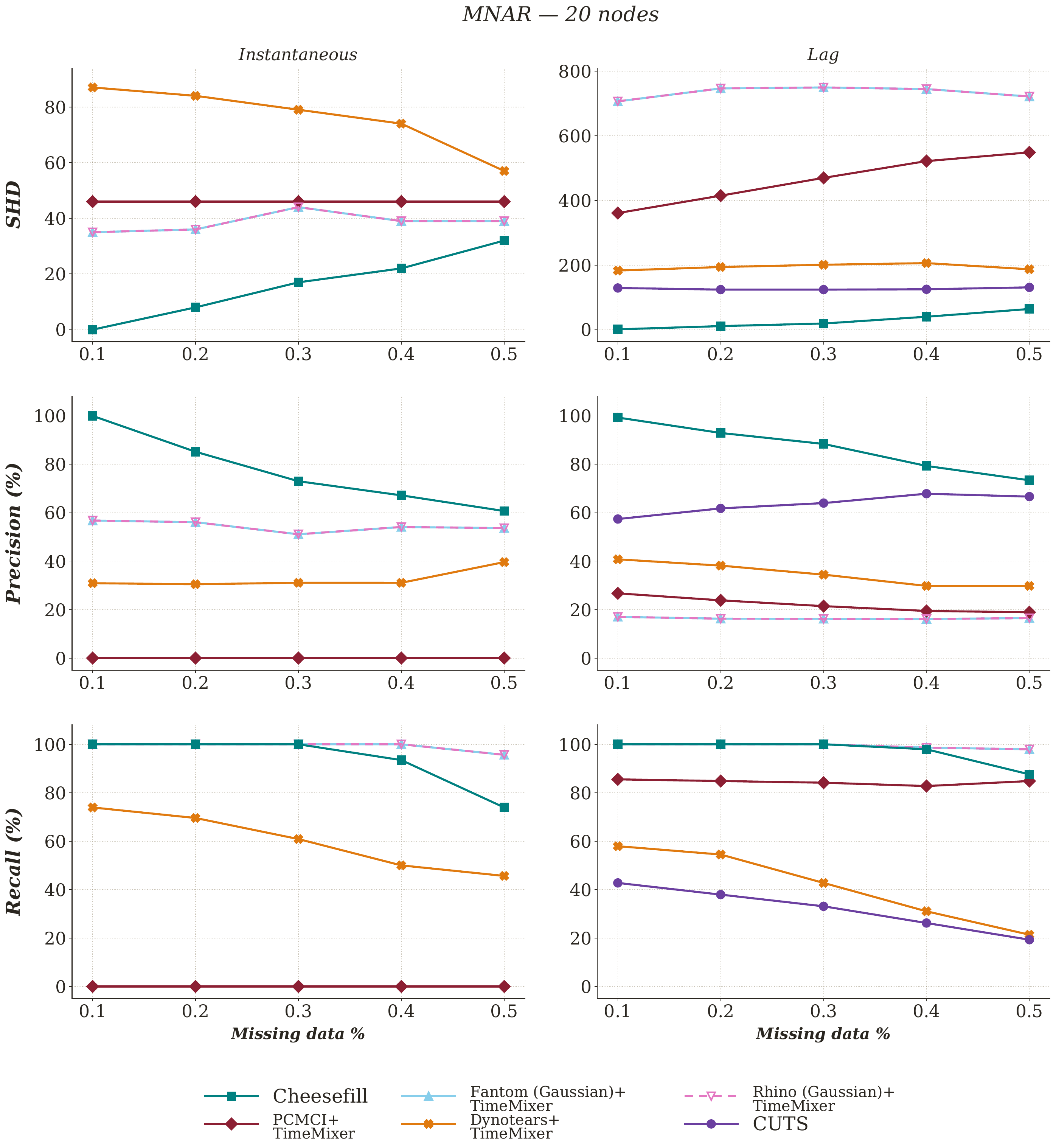}
    \caption{MNAR MTS with $20$ nodes. SHD (top, lower is better), Precision (middle, higher is better) and Recall (bottom, higher is better) for instantaneous and time lagged edges, evaluated under $3$ maximum lags as the missing rate increases from $10\%$ to $50\%$.}
    \label{20_MNAR_synthresult_app_recall}
\end{figure}

\begin{figure}[H]
    \centering
    \includegraphics[width=0.6\linewidth]{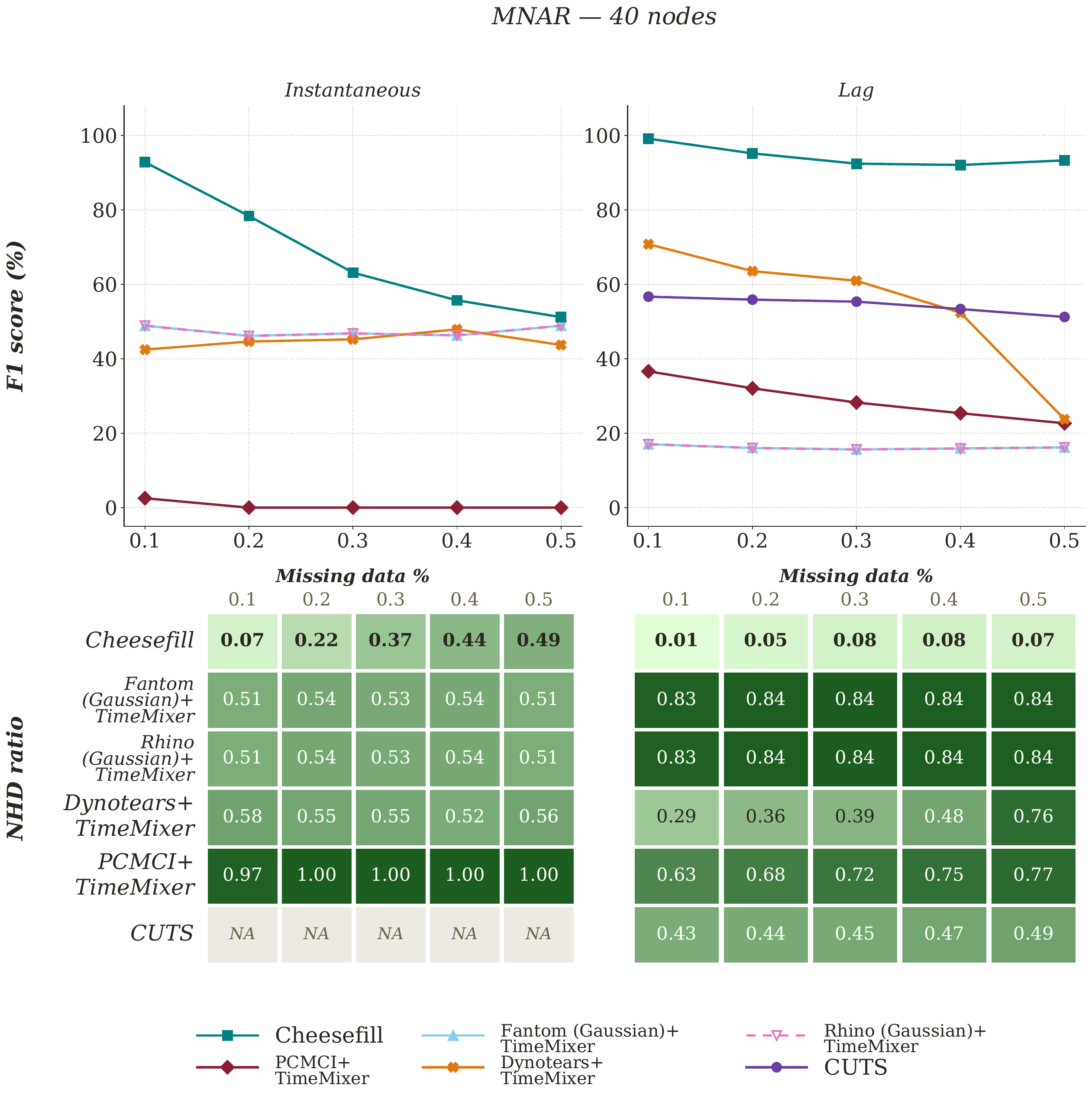}
    \caption{MNAR MTS with $40$ nodes. F1 score (top, higher is better) and NHD ratio (bottom, lower is better) for instantaneous and time lagged edges, evaluated under $3$ maximum lags as the missing rate increases from $10\%$ to $50\%$.}
    \label{40_MNAR_synthresult_app}
\end{figure}

\begin{figure}[H]
    \centering
    \includegraphics[width=0.6\linewidth]{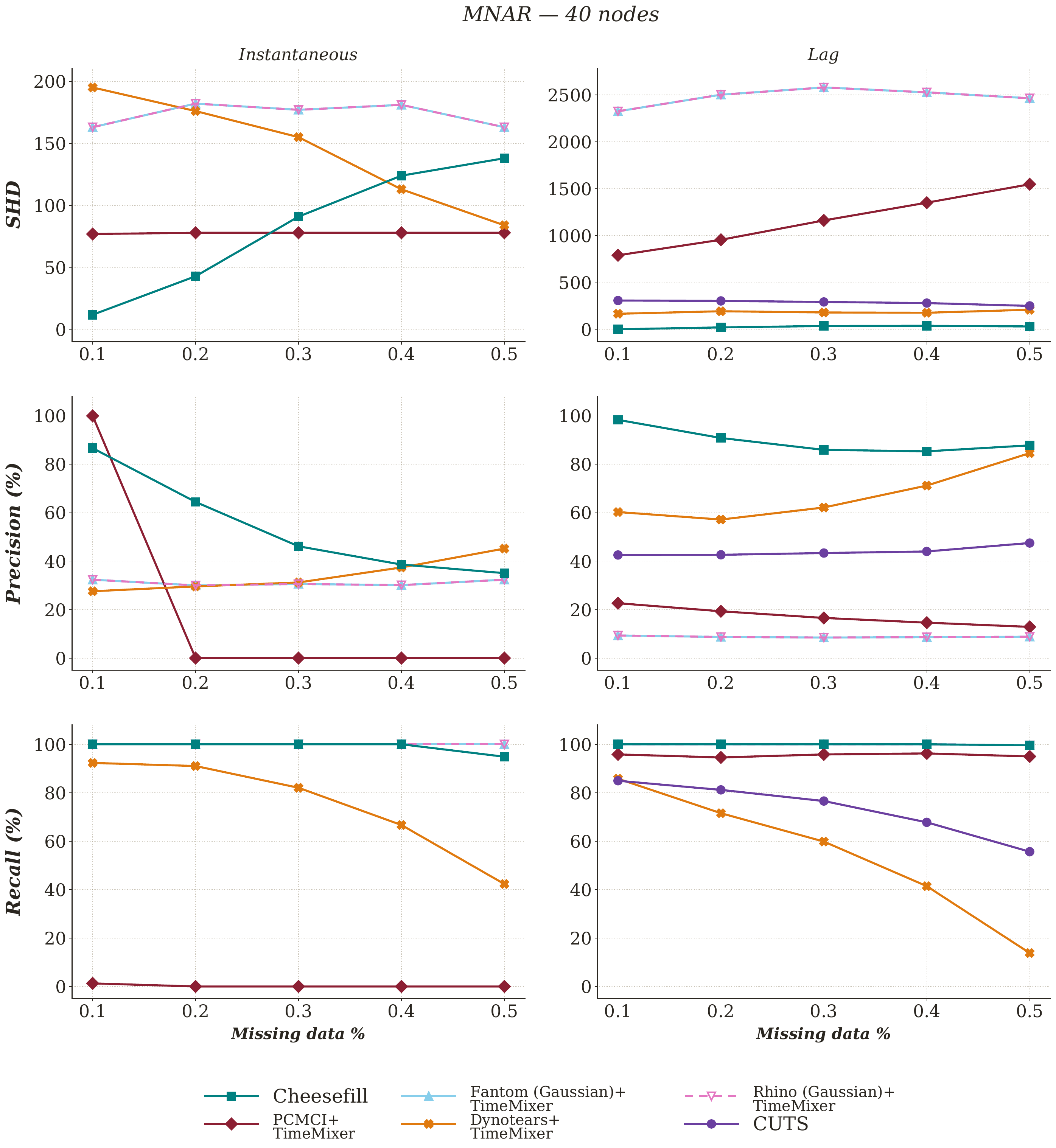}
    \caption{MNAR MTS with $40$ nodes. SHD (top, lower is better), Precision (middle, higher is better) and Recall (bottom, higher is better) for instantaneous and time lagged edges, evaluated under $3$ maximum lags as the missing rate increases from $10\%$ to $50\%$.}
    \label{40_MNAR_synthresult_app_recall}
\end{figure}
\section{Graph average cost details}
\label{app:avg_costtheo_detail}
In this section, we announce the proposition for the graph averaged mechanism cost and we show that the equivalence and parametric upper bound results of Proposition~\ref{thm:correction_ot_equivalence} remain valid when \(c_{\mathcal G}\) is replaced by \(\bar c_\phi\). We present also the full proof for this proposition.
\begin{mybox}
\begin{proposition}[Graph averaged correction map formulation]
\label{thm:graph_average_correction_ot_equivalence}
Let \((\Omega_X,\mathcal B(\Omega_X))\) be the measurable space of complete,
finite length, \(d\) variate MTS trajectories, and assume that \(\Omega_X\) is a
standard Borel space. Let \(P_{\tilde{\boldsymbol X}_{\zeta}}\) be the
pseudo complete data distribution on \(\Omega_X\), and let \(Q_\Theta\) be the
SEM induced complete trajectory distribution on \(\Omega_X\). For a graph
posterior \(q_\phi(\mathcal G)\), define the graph averaged mechanism cost
$\bar c_{\phi}(\tilde{\boldsymbol X},\boldsymbol Y)
:=
\mathbb{E}_{\mathcal G\sim q_\phi(\mathcal G)}
\bigl[
c_{\mathcal G}(\tilde{\boldsymbol X},\boldsymbol Y)
\bigr],$ and assume that \(\bar c_\phi\) is measurable and satisfies
$\mathbb E_{(X,Y)\sim\Gamma}
\bigl[
|\bar c_\phi(X,Y)|
\bigr]<\infty$ for all couplings \(\Gamma\) considered. Let
$\mathcal H(q_\phi)
=
-\mathbb E_{q_\phi}[\log q_\phi],$
and assume \(\mathcal H(q_\phi)<\infty\). Then, for any fixed \(\phi\) and
\(\beta>0\),
$$W_{\bar c_\phi}
\bigl(
P_{\tilde{\boldsymbol X}_{\zeta}},
Q_\Theta
\bigr)
-
\beta\mathcal H(q_\phi)
\nonumber =
\inf_{T\in\mathcal F(P_{\tilde{\boldsymbol X}_{\zeta}},Q_\Theta)}
\left\{
\mathbb{E}_{\tilde{\boldsymbol X}\sim P_{\tilde{\boldsymbol X}_{\zeta}}}
\mathbb{E}_{\boldsymbol Y\sim T(\cdot\mid\tilde{\boldsymbol X})}
\bigl[
\bar c_{\phi}(\tilde{\boldsymbol X},\boldsymbol Y)
\bigr]
-
\beta\mathcal H(q_\phi)
\right\},$$

where \(\mathcal F(P_{\tilde{\boldsymbol X}_{\zeta}},Q_\Theta)\) is the set of
admissible stochastic kernels in Definition~\ref{def:admissible_kernel}.

Moreover, for any parametric family \(\{T_\psi\}_{\psi\in\Psi}\), let
\[
\Psi_{\mathrm{adm}}
=
\{\psi\in\Psi:
T_\psi\in\mathcal F(P_{\tilde{\boldsymbol X}_{\zeta}},Q_\Theta)\}.
\]
Then
$$W_{\bar c_\phi}
\bigl(
P_{\tilde{\boldsymbol X}_{\zeta}},
Q_\Theta
\bigr)
-
\beta\mathcal H(q_\phi)
\nonumber \leq
\inf_{\psi\in\Psi_{\mathrm{adm}}}
\left\{
\mathbb{E}_{\tilde{\boldsymbol X}\sim P_{\tilde{\boldsymbol X}_{\zeta}}}
\mathbb{E}_{\boldsymbol Y\sim T_\psi(\cdot\mid\tilde{\boldsymbol X})}
\bigl[
\bar c_{\phi}(\tilde{\boldsymbol X},\boldsymbol Y)
\bigr]
-
\beta\mathcal H(q_\phi)
\right\}.$$
The inequality is tight whenever the parametric family contains an optimal
admissible kernel for the graph averaged cost \(\bar c_\phi\). More generally,
if the family contains an \(\varepsilon\)-optimal admissible kernel, then the
right-hand side is at most \(\varepsilon\) above the unrestricted optimum.
\end{proposition}
\end{mybox}
\begin{proof}
For readability, write
\[
P := P_{\tilde{\boldsymbol X}_{\zeta}},
\qquad
Q := Q_\Theta .
\]
Both \(P\) and \(Q\) are probability measures on the same measurable trajectory
space \((\Omega_X,\mathcal B(\Omega_X))\).

Fix \(\phi\). Since
\[
\bar c_\phi(x,y)
=
\mathbb E_{\mathcal G\sim q_\phi(\mathcal G)}
[c_{\mathcal G}(x,y)]
\]
is assumed measurable and integrable under the couplings considered,
\(\bar c_\phi\) is a valid transport cost for the Kantorovich problem
\[
W_{\bar c_\phi}(P,Q)
=
\inf_{\Gamma\in\mathcal P(X\sim P,\,Y\sim Q)}
\mathbb E_{(X,Y)\sim\Gamma}
[
\bar c_\phi(X,Y)
].
\]

We first show that this Kantorovich problem is equivalent to an optimization over
admissible stochastic kernels. Take any coupling
\[
\Gamma\in\mathcal P(X\sim P,Y\sim Q).
\]
Since \(\Omega_X\) is a standard Borel space, \(\Gamma\) admits a regular
conditional distribution of the second coordinate given the first. Therefore,
there exists a stochastic kernel \(T_\Gamma(dy\mid x)\) such that
\[
\Gamma(dx,dy)=P(dx)T_\Gamma(dy\mid x).
\]
For every measurable set \(A\in\mathcal B(\Omega_X)\),
\[
\int_{\Omega_X}T_\Gamma(A\mid x)\,dP(x)
=
\Gamma(\Omega_X\times A).
\]
Because \(\Gamma\) has second marginal \(Q\),
\[
\Gamma(\Omega_X\times A)=Q(A).
\]
Thus,
\[
\int_{\Omega_X}T_\Gamma(A\mid x)\,dP(x)=Q(A),
\]
so \(T_\Gamma\in\mathcal F(P,Q)\). Moreover, by disintegration,
\[
\mathbb E_{(X,Y)\sim\Gamma}
[
\bar c_\phi(X,Y)
]
=
\mathbb E_{X\sim P}
\mathbb E_{Y\sim T_\Gamma(\cdot\mid X)}
[
\bar c_\phi(X,Y)
].
\]
Hence every Kantorovich coupling induces an admissible stochastic kernel with
the same graph-averaged mechanism cost.

Conversely, take any admissible stochastic kernel
\[
T\in\mathcal F(P,Q).
\]
Define a probability measure \(\Gamma_T\) on
\(\Omega_X\times\Omega_X\) by
\[
\Gamma_T(dx,dy)=P(dx)T(dy\mid x).
\]
Equivalently, for measurable rectangles \(B\times A\),
\[
\Gamma_T(B\times A)=\int_B T(A\mid x)\,dP(x).
\]
For any \(B\in\mathcal B(\Omega_X)\),
\[
\Gamma_T(B\times\Omega_X)
=
\int_B T(\Omega_X\mid x)\,dP(x)
=
P(B),
\]
so the first marginal of \(\Gamma_T\) is \(P\). For any
\(A\in\mathcal B(\Omega_X)\),
\[
\Gamma_T(\Omega_X\times A)
=
\int_{\Omega_X}T(A\mid x)\,dP(x).
\]
Since \(T\) is admissible, the right-hand side equals \(Q(A)\). Therefore the
second marginal of \(\Gamma_T\) is \(Q\), and
\[
\Gamma_T\in\mathcal P(X\sim P,Y\sim Q).
\]
The cost of this coupling is
\[
\mathbb E_{(X,Y)\sim\Gamma_T}
[
\bar c_\phi(X,Y)
]
=
\mathbb E_{X\sim P}
\mathbb E_{Y\sim T(\cdot\mid X)}
[
\bar c_\phi(X,Y)
].
\]

Thus, every coupling induces an admissible kernel with the same
\(\bar c_\phi\)-cost, and every admissible kernel induces a coupling with the
same \(\bar c_\phi\)-cost. Hence
\[
W_{\bar c_\phi}(P,Q)
=
\inf_{T\in\mathcal F(P,Q)}
\mathbb E_{X\sim P}
\mathbb E_{Y\sim T(\cdot\mid X)}
[
\bar c_\phi(X,Y)
].
\]

Now consider the entropy term. For fixed \(\phi\), the quantity
\[
-\beta\mathcal H(q_\phi)
\]
does not depend on the coupling \(\Gamma\), the admissible kernel \(T\), or the
parametric correction map \(T_\psi\). It is therefore an additive constant with
respect to the transport optimization. Subtracting this same constant from both
sides gives
\[
W_{\bar c_\phi}(P,Q)-\beta\mathcal H(q_\phi)
=
\inf_{T\in\mathcal F(P,Q)}
\left\{
\mathbb E_{X\sim P}
\mathbb E_{Y\sim T(\cdot\mid X)}
[
\bar c_\phi(X,Y)
]
-
\beta\mathcal H(q_\phi)
\right\},
\]
which proves the equality.
Substituting back
\(P=P_{\tilde{\boldsymbol X}_{\zeta}}\) and \(Q=Q_\Theta\) gives the stated
form.

It remains to prove the parametric upper bound. By definition,
\[
\{T_\psi:\psi\in\Psi_{\mathrm{adm}}\}
\subseteq
\mathcal F(P,Q).
\]
Therefore, taking the infimum over the restricted parametric family cannot give
a smaller value than taking the infimum over all admissible kernels:
\[
\inf_{T\in\mathcal F(P,Q)}
\mathbb E_{X\sim P}
\mathbb E_{Y\sim T(\cdot\mid X)}
[
\bar c_\phi(X,Y)
]
\leq
\inf_{\psi\in\Psi_{\mathrm{adm}}}
\mathbb E_{X\sim P}
\mathbb E_{Y\sim T_\psi(\cdot\mid X)}
[
\bar c_\phi(X,Y)
].
\]
Again, since \(-\beta\mathcal H(q_\phi)\) is independent of \(T\) and
\(\psi\), subtracting it from both sides preserves the inequality:
\begin{align*}
&
W_{\bar c_\phi}(P,Q)
-
\beta\mathcal H(q_\phi)
\\
&\qquad \leq
\inf_{\psi\in\Psi_{\mathrm{adm}}}
\left\{
\mathbb E_{X\sim P}
\mathbb E_{Y\sim T_\psi(\cdot\mid X)}
[
\bar c_\phi(X,Y)
]
-
\beta\mathcal H(q_\phi)
\right\}.
\end{align*}
This proves the inequality.

If the parametric family contains an optimal admissible kernel for
\(\bar c_\phi\), then the restricted parametric infimum equals the unrestricted
kernel infimum, so the inequality is tight.

More generally, suppose there exists
\(\psi_\varepsilon\in\Psi_{\mathrm{adm}}\) such that
\[
\mathbb E_{X\sim P}
\mathbb E_{Y\sim T_{\psi_\varepsilon}(\cdot\mid X)}
[
\bar c_\phi(X,Y)
]
\leq
W_{\bar c_\phi}(P,Q)+\varepsilon.
\]
Since the entropy term is the same constant on both sides,
\[
\mathbb E_{X\sim P}
\mathbb E_{Y\sim T_{\psi_\varepsilon}(\cdot\mid X)}
[
\bar c_\phi(X,Y)
]
-
\beta\mathcal H(q_\phi)
\leq
W_{\bar c_\phi}(P,Q)
-
\beta\mathcal H(q_\phi)
+
\varepsilon.
\]
Therefore the restricted parametric objective is at most \(\varepsilon\) above
the unrestricted graph-averaged Kantorovich objective with entropy
regularization.
\end{proof}
\section{Proof of our proposition}
\subsection{Assumptions}
In this section, we define and clarify the required assumptions announced in the main text for Cheesefill.
\label{app_assumptions}
 \begin{assumption}[Causal Stationarity, \citep{runge2018causal}]
    A stationary time series process $(\boldsymbol{x}_t)_{t \in \mathcal{T}}$ with graph $\mathcal{G}$ is called causally stationary over a time index set $\mathcal{T}$ if and only if for all links $\boldsymbol{x}_{t-\tau}^i \rightarrow \boldsymbol{x}_t^j$ in the graph
\begin{equation*}
    x_{t-\tau}^i \not\!\perp\!\!\!\perp x_t^j \mid \boldsymbol{x}_{<t} \setminus \{x_{t-\tau}^i\}.
 \vspace{-10pt}   
\end{equation*}
\label{def_caus_sta}
\end{assumption}
This elucidates the inherent characteristics of the time-series data generation mechanism, thereby validating the choice of the auto-regressive model.

\begin{assumption} (Causal Markov Property, \cite{peters2017elements}). Given a DAG $\mathcal{G}$ and a joint distribution $p$, this distribution is said to satisfy causal Markov property w.r.t. the DAG $\mathcal{G}$ if each variable is independent of its non-descendants given its parents. 
\label{markovprop}
\end{assumption}
This is a common assumptions for the distribution induced by an SEM. With this assumption, one
can deduce conditional independence between variables from the graph.
\begin{assumption}[Causal Markov Property (CMP)]
A joint distributions $(p(\cdot|\mathcal{G}))$ satisfies the \textbf{CMP} with respect to the DAG $(\mathcal{G})$ if each variable is independent of its non-descendants given its parents.
\label{mainass2}
\end{assumption}
\vspace{-2pt}
\begin{assumption}[Causal Minimality]
 Given DAG $(\mathcal{G})$ and a joint distribution $(p(\cdot|\mathcal{G}))$, we say that this distributions satisfies causal minimality w.r.t. the DAG $(\mathcal{G})$ if $p(\cdot|\mathcal{G})$ is Markovian w.r.t the DAG $\mathcal{G}$ but not to any proper subgraph of $\mathcal{G}$.
\label{mainass3}
\end{assumption}
\vspace{-2pt}
\begin{assumption}[Causal Sufficiency]
A set of observed variables $\boldsymbol{V}$ is causally sufficient for a process $\boldsymbol{x}_t$ if and only if in the process every common cause of any two or more variables in $\boldsymbol{V}$ is in $\boldsymbol{V}$ or has the same value for all units in the population.  \label{mainass4}
\end{assumption}
This assumption implies there are no latent confounders present in the time-series data.
\subsection{Proof}
\label{app_proof}
\begin{mybox}\begin{proposition}[]
Let \((\Omega_X,\mathcal B(\Omega_X))\) be the measurable space of complete
length \(|\mathcal{T}|\), \(d\) variate MTS trajectories. Assume that \(\Omega_X\) is a
standard Borel space. Let \(P_{\tilde{\boldsymbol X}_{\zeta}}\) be the
pseudo complete data distribution on \(\Omega_X\), and let \(Q_\Theta\) be the
SEM induced complete trajectory distribution on \(\Omega_X\). Let \(c_{\mathcal G}:\Omega_X\times\Omega_X\to\mathbb R_+\) be a measurable
mechanism cost satisfying $\mathbb E_{(X,Y)\sim\Gamma}\bigl[|c_{\mathcal G}(X,Y)|\bigr]<\infty$
for all couplings \(\Gamma\) considered. Then
\begin{equation}
W_{c_{\mathcal G}}
\bigl(
P_{\tilde{\boldsymbol X}_{\zeta}},
Q_\Theta
\bigr)
=
\inf_{T\in\mathcal F(P_{\tilde{\boldsymbol X}_{\zeta}},Q_\Theta)}
\mathbb{E}_{\tilde{\boldsymbol X}\sim P_{\tilde{\boldsymbol X}_{\zeta}}}
\mathbb{E}_{\boldsymbol Y\sim T(\cdot\mid\tilde{\boldsymbol X})}
\bigl[
c_{\mathcal G}(\tilde{\boldsymbol X},\boldsymbol Y)
\bigr],
\end{equation}
where \(\mathcal F(P_{\tilde{\boldsymbol X}_{\zeta}},Q_\Theta)\) is the set of
admissible stochastic kernels in Definition~\ref{def:admissible_kernel}.
Moreover, for any parametric family \(\{T_\psi\}_{\psi\in\Psi}\), let $\Psi_{\mathrm{adm}}
=
\{\psi\in\Psi:
T_\psi\in\mathcal F(P_{\tilde{\boldsymbol X}_{\zeta}},Q_\Theta)\}.$ Then
\begin{equation}
W_{c_{\mathcal G}}
\bigl(
P_{\tilde{\boldsymbol X}_{\zeta}},
Q_\Theta
\bigr)
\leq
\inf_{\psi\in\Psi_{\mathrm{adm}}}
\mathbb{E}_{\tilde{\boldsymbol X}\sim P_{\tilde{\boldsymbol X}_{\zeta}}}
\mathbb{E}_{\boldsymbol Y\sim T_\psi(\cdot\mid\tilde{\boldsymbol X})}
\bigl[
c_{\mathcal G}(\tilde{\boldsymbol X},\boldsymbol Y)
\bigr].
\end{equation}
The inequality is tight whenever the parametric family contains an optimal
admissible kernel. More generally, if the family contains an
\(\varepsilon\)-optimal admissible kernel, then the right-hand side is at most
\(\varepsilon\) above the unrestricted Kantorovich optimum.
\end{proposition}\end{mybox}
\begin{proof}
For readability, write
\[
P := P_{\tilde{\boldsymbol X}_{\zeta}},
\qquad
Q := Q_\Theta .
\]
Both \(P\) and \(Q\) are probability measures on the same measurable trajectory
space \((\Omega_X,\mathcal B(\Omega_X))\). By definition, the Kantorovich cost is
\[
W_{c_{\mathcal G}}(P,Q)
=
\inf_{\Gamma\in\mathcal P(X\sim P,\,Y\sim Q)}
\mathbb E_{(X,Y)\sim\Gamma}
\bigl[
c_{\mathcal G}(X,Y)
\bigr],
\]
where \(\mathcal P(X\sim P,Y\sim Q)\) denotes the set of couplings whose first
marginal is \(P\) and whose second marginal is \(Q\).

We prove that optimizing over couplings
\(\Gamma\in\mathcal P(X\sim P,Y\sim Q)\) is equivalent to optimizing over
admissible stochastic kernels \(T\in\mathcal F(P,Q)\).

First, take any coupling
\[
\Gamma\in\mathcal P(X\sim P,Y\sim Q).
\]
Since \(\Omega_X\) is a standard Borel space, every probability measure on
\(\Omega_X\times\Omega_X\) admits a regular conditional distribution. Therefore,
there exists a stochastic kernel \(T_\Gamma(dy\mid x)\) such that
\[
\Gamma(dx,dy)=P(dx)T_\Gamma(dy\mid x).
\]
We now show that \(T_\Gamma\) is admissible. For any measurable set
\(A\in\mathcal B(\Omega_X)\),
\[
\int_{\Omega_X}T_\Gamma(A\mid x)\,dP(x)
=
\Gamma(\Omega_X\times A).
\]
Because \(\Gamma\) has second marginal \(Q\), we have
\[
\Gamma(\Omega_X\times A)=Q(A).
\]
Hence
\[
\int_{\Omega_X}T_\Gamma(A\mid x)\,dP(x)=Q(A),
\]
so \(T_\Gamma\in\mathcal F(P,Q)\). Moreover, by the disintegration identity,
\[
\mathbb E_{(X,Y)\sim\Gamma}
\bigl[
c_{\mathcal G}(X,Y)
\bigr]
=
\mathbb E_{X\sim P}
\mathbb E_{Y\sim T_\Gamma(\cdot\mid X)}
\bigl[
c_{\mathcal G}(X,Y)
\bigr].
\]
The equality is well-defined by the assumed measurability and integrability of
\(c_{\mathcal G}\). Therefore, every Kantorovich coupling induces an admissible
stochastic kernel with exactly the same cost.

Conversely, take any admissible stochastic kernel
\[
T\in\mathcal F(P,Q).
\]
Define a probability measure \(\Gamma_T\) on
\(\Omega_X\times\Omega_X\) by
\[
\Gamma_T(dx,dy)=P(dx)T(dy\mid x).
\]
Equivalently, for measurable rectangles
\(B\times A\), with \(A,B\in\mathcal B(\Omega_X)\),
\[
\Gamma_T(B\times A)
=
\int_B T(A\mid x)\,dP(x).
\]
We verify that \(\Gamma_T\) is a coupling of \(P\) and \(Q\). For any measurable
set \(B\in\mathcal B(\Omega_X)\),
\[
\Gamma_T(B\times\Omega_X)
=
\int_B T(\Omega_X\mid x)\,dP(x)
=
\int_B 1\,dP(x)
=
P(B),
\]
so the first marginal of \(\Gamma_T\) is \(P\). Similarly, for any measurable
set \(A\in\mathcal B(\Omega_X)\),
\[
\Gamma_T(\Omega_X\times A)
=
\int_{\Omega_X}T(A\mid x)\,dP(x).
\]
Since \(T\) is admissible, the right-hand side equals \(Q(A)\). Therefore,
\[
\Gamma_T(\Omega_X\times A)=Q(A),
\]
so the second marginal of \(\Gamma_T\) is \(Q\). Hence
\[
\Gamma_T\in\mathcal P(X\sim P,Y\sim Q).
\]
The cost of this coupling is
\[
\mathbb E_{(X,Y)\sim\Gamma_T}
\bigl[
c_{\mathcal G}(X,Y)
\bigr]
=
\mathbb E_{X\sim P}
\mathbb E_{Y\sim T(\cdot\mid X)}
\bigl[
c_{\mathcal G}(X,Y)
\bigr].
\]
Thus, every admissible stochastic kernel induces a Kantorovich coupling with
exactly the same cost.

The two constructions show that the optimization over Kantorovich couplings and
the optimization over admissible stochastic kernels have the same feasible
objects up to disintegration, and the same objective value. Therefore,
\[
W_{c_{\mathcal G}}(P,Q)
=
\inf_{T\in\mathcal F(P,Q)}
\mathbb E_{X\sim P}
\mathbb E_{Y\sim T(\cdot\mid X)}
\bigl[
c_{\mathcal G}(X,Y)
\bigr].
\]
Substituting back
\[
P=P_{\tilde{\boldsymbol X}_{\zeta}},
\qquad
Q=Q_\Theta,
\]
gives Eq.~\eqref{eq:correction_ot_equivalence}.

It remains to prove the parametric upper bound. By definition,
\[
\Psi_{\mathrm{adm}}
=
\{\psi\in\Psi:T_\psi\in\mathcal F(P,Q)\}.
\]
Hence the parametric admissible family
\[
\{T_\psi:\psi\in\Psi_{\mathrm{adm}}\}
\]
is a subset of the unrestricted admissible kernel class \(\mathcal F(P,Q)\).
Taking an infimum over a smaller feasible set cannot produce a smaller value
than taking an infimum over the full feasible set. Therefore,
\[
\inf_{T\in\mathcal F(P,Q)}
\mathbb E_{X\sim P}
\mathbb E_{Y\sim T(\cdot\mid X)}
\bigl[
c_{\mathcal G}(X,Y)
\bigr]
\leq
\inf_{\psi\in\Psi_{\mathrm{adm}}}
\mathbb E_{X\sim P}
\mathbb E_{Y\sim T_\psi(\cdot\mid X)}
\bigl[
c_{\mathcal G}(X,Y)
\bigr].
\]
Using the equality already proved, the left-hand side is
\(W_{c_{\mathcal G}}(P,Q)\). Hence
\[
W_{c_{\mathcal G}}(P,Q)
\leq
\inf_{\psi\in\Psi_{\mathrm{adm}}}
\mathbb E_{X\sim P}
\mathbb E_{Y\sim T_\psi(\cdot\mid X)}
\bigl[
c_{\mathcal G}(X,Y)
\bigr],
\]
which proves Eq.~\eqref{eq:parametric_upper_bound}. Substituting back
\(P=P_{\tilde{\boldsymbol X}_{\zeta}}\) and \(Q=Q_\Theta\) gives the stated
bound.

If the parametric family contains an optimal admissible kernel, then there
exists \(\psi^\star\in\Psi_{\mathrm{adm}}\) such that
\[
T_{\psi^\star}
\in
\arg\min_{T\in\mathcal F(P,Q)}
\mathbb E_{X\sim P}
\mathbb E_{Y\sim T(\cdot\mid X)}
\bigl[
c_{\mathcal G}(X,Y)
\bigr].
\]
In that case, the restricted parametric infimum attains the same value as the
unrestricted infimum, and the inequality in
Eq.~\eqref{eq:parametric_upper_bound} is tight.

More generally, suppose the parametric family contains an
\(\varepsilon\)-optimal admissible kernel. That is, suppose there exists
\(\psi_\varepsilon\in\Psi_{\mathrm{adm}}\) such that
\[
\mathbb E_{X\sim P}
\mathbb E_{Y\sim T_{\psi_\varepsilon}(\cdot\mid X)}
\bigl[
c_{\mathcal G}(X,Y)
\bigr]
\leq
W_{c_{\mathcal G}}(P,Q)+\varepsilon.
\]
Then, since \(T_{\psi_\varepsilon}\) is one feasible element of the restricted
parametric class,
\[
\inf_{\psi\in\Psi_{\mathrm{adm}}}
\mathbb E_{X\sim P}
\mathbb E_{Y\sim T_\psi(\cdot\mid X)}
\bigl[
c_{\mathcal G}(X,Y)
\bigr]
\leq
W_{c_{\mathcal G}}(P,Q)+\varepsilon.
\]
Combining this with the lower bound just proved gives
\[
W_{c_{\mathcal G}}(P,Q)
\leq
\inf_{\psi\in\Psi_{\mathrm{adm}}}
\mathbb E_{X\sim P}
\mathbb E_{Y\sim T_\psi(\cdot\mid X)}
\bigl[
c_{\mathcal G}(X,Y)
\bigr]
\leq
W_{c_{\mathcal G}}(P,Q)+\varepsilon.
\]
Therefore the parametric objective is at most \(\varepsilon\) above the
unrestricted Kantorovich optimum whenever the parametric family contains an
\(\varepsilon\)-optimal admissible kernel.
\end{proof}

Proposition~\ref{thm:correction_ot_equivalence} shows that the correction map is
a conditional representation of an OT coupling. A coupling
\(\Gamma\in\mathcal P(X\sim P_{\tilde{\boldsymbol X}_{\zeta}},
Y\sim Q_\Theta)\) specifies how mass from the pseudo complete distribution is
matched to mass from the SEM induced distribution. Since the trajectory space is
standard Borel, this coupling can be disintegrated into the source marginal
\(P_{\tilde{\boldsymbol X}_{\zeta}}\) and a conditional kernel
\(T(d\boldsymbol Y\mid\tilde{\boldsymbol X})\). Conversely, any admissible
kernel defines a coupling with the required marginals. Thus, the stochastic
correction map is not an auxiliary imputation heuristic; it is an amortized
parameterization of the conditional component of a Kantorovich coupling.

\newpage
\section*{NeurIPS Paper Checklist}
\begin{enumerate}

\item {\bf Claims}
    \item[] Question: Do the main claims made in the abstract and introduction accurately reflect the paper's contributions and scope?
    \item[] Answer: \answerYes{} 
    \item[] Justification: Yes the main claims are clear in the abstract, introduction and in the whole paper

\item {\bf Limitations}
    \item[] Question: Does the paper discuss the limitations of the work performed by the authors?
    \item[] Answer: \answerYes{} 
    \item[] Justification: Yes, we have a section in appendix \ref{limitations} in which we talk about the limitation of our work.

\item {\bf Theory assumptions and proofs}
    \item[] Question: For each theoretical result, does the paper provide the full set of assumptions and a complete (and correct) proof?
    \item[] Answer: \answerYes{} 
    \item[] Justification: All assumptions and complete proofs appear in Appendix \ref{app_proof}. 

    \item {\bf Experimental result reproducibility}
    \item[] Question: Does the paper fully disclose all the information needed to reproduce the main experimental results of the paper to the extent that it affects the main claims and/or conclusions of the paper (regardless of whether the code and data are provided or not)?
    \item[] Answer: \answerYes{} 
    \item[] Justification: Yes, we provide in the main text figures that illustrate the exact architecture used for our model. In the appendices, we provide the exact hyper-parameters needed to reproduce our results in all the presented experiments. \textcolor{teal}{Our code is also provided in the supplementary material}. 

\item {\bf Open access to data and code}
    \item[] Question: Does the paper provide open access to the data and code, with sufficient instructions to faithfully reproduce the main experimental results, as described in supplemental material?
    \item[] Answer: \answerYes{} 
    \item[] Justification: Our code, with all the instruction to reproduce our results, is provided in the supplementary materials 

\item {\bf Experimental setting/details}
    \item[] Question: Does the paper specify all the training and test details (e.g., data splits, hyperparameters, how they were chosen, type of optimizer) necessary to understand the results?
    \item[] Answer: \answerYes{} 
    \item[] Justification: All the detailed needed to run our code and to reproduce our results are presented in Appendix \ref{optim_param}.

\item {\bf Experiment statistical significance}
    \item[] Question: Does the paper report error bars suitably and correctly defined or other appropriate information about the statistical significance of the experiments?
    \item[] Answer: \answerYes{} 
    \item[] Justification: We provide results with standard deviations in all our experiments. 

\item {\bf Experiments compute resources}
    \item[] Question: For each experiment, does the paper provide sufficient information on the computer resources (type of compute workers, memory, time of execution) needed to reproduce the experiments?
    \item[] Answer: \answerYes{}
    \item[] Justification: Our model takes 4 minutes to run on A100 with 80 Gb
    
\item {\bf Code of ethics}
    \item[] Question: Does the research conducted in the paper conform, in every respect, with the NeurIPS Code of Ethics \url{https://neurips.cc/public/EthicsGuidelines}?
    \item[] Answer: \answerYes{}
    \item[] Justification: -

\item {\bf Broader impacts}
    \item[] Question: Does the paper discuss both potential positive societal impacts and negative societal impacts of the work performed?
    \item[] Answer: \answerYes{} 
    \item[] Justification: In our introduction, we talk about paper’s broader impact.
    
\item {\bf Safeguards}
    \item[] Question: Does the paper describe safeguards that have been put in place for responsible release of data or models that have a high risk for misuse (e.g., pre-trained language models, image generators, or scraped datasets)?
    \item[] Answer: \answerNA{}  
    \item[] Justification: -

\item {\bf Licenses for existing assets}
    \item[] Question: Are the creators or original owners of assets (e.g., code, data, models), used in the paper, properly credited and are the license and terms of use explicitly mentioned and properly respected?
    \item[] Answer: \answerYes{} 
    \item[] Justification: -

\item {\bf New assets}
    \item[] Question: Are new assets introduced in the paper well documented and is the documentation provided alongside the assets?
    \item[] Answer: \answerYes{}
    \item[] Justification: e presented all the details about our model: parameters, architecture, data.
Also the detailed proofs are provided

\item {\bf Crowdsourcing and research with human subjects}
    \item[] Question: For crowdsourcing experiments and research with human subjects, does the paper include the full text of instructions given to participants and screenshots, if applicable, as well as details about compensation (if any)? 
    \item[] Answer: \answerNA{}
    \item[] Justification: -

\item {\bf Institutional review board (IRB) approvals or equivalent for research with human subjects}
    \item[] Question: Does the paper describe potential risks incurred by study participants, whether such risks were disclosed to the subjects, and whether Institutional Review Board (IRB) approvals (or an equivalent approval/review based on the requirements of your country or institution) were obtained?
    \item[] Answer: \answerNA{}
    \item[] Justification: -

\item {\bf Declaration of LLM usage}
    \item[] Question: Does the paper describe the usage of LLMs if it is an important, original, or non-standard component of the core methods in this research? Note that if the LLM is used only for writing, editing, or formatting purposes and does \emph{not} impact the core methodology, scientific rigor, or originality of the research, declaration is not required.
    \item[] Answer: \answerNA{} 
    \item[] Justification: -

\end{enumerate}

\end{document}